\documentclass[10pt,letterpaper]{article}

\usepackage[letterpaper,textwidth=5.5in,textheight=9in,heightrounded]{geometry}
\usepackage[utf8]{inputenc}
\usepackage[T1]{fontenc}
\usepackage{times}
\usepackage{microtype}

\usepackage{abstract}
\usepackage{amsmath,amssymb,amsfonts,amsthm}
\usepackage{stmaryrd}
\usepackage{nicefrac}

\usepackage{graphicx}
\usepackage{tikz}
\usetikzlibrary{arrows.meta,positioning,decorations.pathreplacing,calc}
\usepackage{enumitem}
\usepackage{booktabs}
\usepackage{xcolor}

\usepackage{natbib}
\usepackage{url}
\usepackage{hyperref}
\hypersetup{colorlinks=true,linkcolor=black,citecolor=black,urlcolor=black}
\usepackage[capitalize]{cleveref}

\usepackage{titlesec}
\titleformat{\section}{\large\bfseries}{\thesection}{1em}{}
\titleformat{\subsection}{\normalsize\bfseries}{\thesubsection}{1em}{}
\titlespacing*{\section}{0pt}{1.5ex plus 1ex minus .2ex}{0.8ex plus .2ex}
\titlespacing*{\subsection}{0pt}{1.2ex plus 1ex minus .2ex}{0.5ex plus .2ex}
\titlespacing*{\subsubsection}{0pt}{1ex plus 1ex minus .2ex}{0.3ex plus .2ex}
\titlespacing*{\paragraph}{0pt}{1ex plus .5ex minus .2ex}{1em}

\newtheorem{theorem}{Theorem}

\newtheorem{proposition}{Proposition}

\newtheorem{definition}{Definition}

\theoremstyle{definition}
\newtheorem{remark}{Remark}
\newtheorem{openproblem}{Open Problem}

\newcommand{\Prf}{\mathrm{Prf}}
\newcommand{\Test}{\mathrm{Test}}
\newcommand{\pw}[1]{\frac{1\phantom{{}^{#1}}}{2^{#1}}}   
\newcommand{\tpw}[1]{1/2^{#1}}                          

\title{\bfseries Machine Learning Beyond Statistical Consistency for Hybrid AI}

\author{%
  \normalsize\setlength{\tabcolsep}{0pt}%
  \begin{tabular}[t]{c}
    Jinsook Kim \\
    Korea Institute for Advanced Study \\
    Seoul, Korea \\
    \texttt{jki76364@gmail.com}
  \end{tabular}\hspace{2.5em}%
  \begin{tabular}[t]{c}
    Jinho Kang \\
    Gwangju Institute of Science and Technology \\
    Gwangju, Korea \\
    \texttt{jhkang919@gmail.com}
  \end{tabular}%
}
\date{}

\begin{document}

\maketitle

\begin{abstract}
We propose a new definition of machine learning, according to which a machine learns a function when it computes the function successfully, that is, when the value it returns for each input has a proper derivation in a formal system. By G\"{o}del's Representability Theorem, a machine computes a value exactly when a formal system of arithmetic derives it from the axioms and the mathematical definitions. We extend this equivalence between computation and derivation to empirical targets in two ways: the theorem is extended to the computable reals, and the data are admitted into the system as empirical definitions that record what has been observed. A derivation is proper when each of its premises is an axiom, a mathematical or empirical definition, or a formula derivable from these, and none of them is a global assumption about the data. We prove that a machine learns a true probability if and only if two conditions are met. The Truth Condition is that its estimate converges to the true probability, which is statistical consistency. The Assurance Condition is that the machine properly derives its own convergence inside the system. Learning thus lies beyond statistical consistency, for it requires a derivation of the convergence from the record. A global assumption about the data, such as the i.i.d.\ assumption, does not supply such a derivation, because it leads only to a circular one, as we show for PAC learning. On the other hand, the two conditions are met for the probability of a sentence over a finite ideal corpus. A real machine can approximate a true probability only if an ideal machine can learn it precisely, that is, only in a domain that meets the two conditions. Modern AI works in such a domain, and we offer this as a clue to its success. Modern AI is hybrid as well, in two senses: First, it performs two kinds of tasks, namely probabilistic ones such as predicting the next word and fundamentally deductive ones such as proving and coding. Second, it combines two parts of different kinds, a statistical proposer and a deductive verifier, for proving and coding. Our framework explains both senses of hybridity in one formal system.
\end{abstract}


\section{Introduction}
\label{sec:intro}

Machine learning, according to \citet{JorMit:15}, addresses the question of how to build computer systems that improve automatically through experience. So a ``machine'' here means a \textit{computing} machine. What it is for a machine to compute is traditionally well settled: a machine computes a function when a Turing machine computes it \citep{Turing:36}. By G\"{o}del's Representability Theorem, then, computation is equivalent to instance-wise \textit{derivation}: a Turing machine computes a value of a function if and only if that value is derivable in a formal system of arithmetic \citep{Godel:31}. So a machine computes a function when the formal system derives it value by value. If we now read derivability or provability epistemically, as G\"{o}del and Feferman did \citep{Godel:46, Fefer:06}, then we can say that what the system derives is what the machine \textit{learns}. That is, when a machine derives how the output of a function is obtained from the input, we can take the machine to learn the value of the function. 

Along this line, we define machine learning based on the notion of computation. Our definition is inspired by the three ideas: Turing's analysis of computation, G\"{o}del's theorem that computation and derivation coincide, and the epistemic reading of provability as learning. Up to this point everything is classical mathematics and theoretical computer science. But modern machine learning goes beyond the computation of purely mathematical functions: it computes \textit{empirical} functions, in particular probabilities, from data. So we extend the classical notion of computability to the domain of empirical learning from data, and on that domain we define modern machine learning anew as \textit{successful} computation. 

Our extension starts from the fact that computing a function is deriving it in a formal system. It then takes two steps, one a theorem and one a definition. First, we extend  G\"{o}del's Representability Theorem to the \textit{computable reals}, so that a function such as a probability whose values are real numbers can be the object of a derivation (Theorem~\ref{thm:rep}). Second, we bring the data, which are empirical entities, into the syntactic system as \textit{empirical} definitions that record what has been observed up to time~$t$, and we define the computation of a function as successful when each of its values has a proper derivation in the extended system (Definitions~\ref{def:proper} and~\ref{def:success}). A derivation is \textit{proper} when each of its premises is an axiom, a mathematical or an empirical definition, or what follows from them; it is a \textit{mere} derivation when one of its premises is a global assumption about the data, admitted as though it were an axiom. For a purely mathematical function, the distinction between proper and mere derivations is irrelevant: the values of such a function are fixed by the axioms and the mathematical definitions, so they are derivable from these alone, and such a derivation is proper. For an empirical function, however, it is not so. The values of an empirical function are fixed by the world, so a derivation of a value of the target needs a premise about the world, and that premise is either the record or a global assumption about the data. This is where the distinction does its work, and it is why this paper is about empirical learning.

Successful computation, which defines learning, is to be distinguished from \textit{mere} computation, in which the value of an empirical function is derived from a global assumption about the structure of the data. A global assumption is not derivable from any finite record, so it can enter the system only as an axiom, and a derivation from it is a mere derivation even when the record is used in computing the estimate. The empirical targets of this paper are true probabilities of data-generating processes, and for them a derivation from a global assumption is \textit{circular}. That is, a global assumption about the structure of the data, such as the i.i.d.\ assumption, takes as a premise the very structure that the true probability is, so it already presupposes what the derivation establishes about the target. We therefore hold that a machine that learns a true probability must go beyond derivation from global assumptions. In this respect, our theory of learning differs from existing statistical learning theory, which is based on such assumptions.

We give two conditions under which a machine learns a probability function, that is, computes it successfully without a global assumption. The first is the \textit{Truth Condition}, that the machine's estimate converges to the true probability. The second is the \textit{Assurance Condition}, that the machine properly derives its own convergence inside the system. The second condition asks for a derivation and not for a check, because convergence is a statement about all future times and the record at any time is finite: no finite record can confirm it one observation at a time, and the only finite object that settles such a statement is a derivation. We prove that the two conditions are necessary and sufficient for learning a probability function. We then show a case in natural language processing in which both conditions are met and a true probability is learned.

One point needs care. The Truth Condition describes \textit{precise} learning of a probability function to every precision, whereas modern machine learning aims at practical \textit{approximation}. We make clear how successful learning connects to that practice as follows: where precise learning is not possible, approximation is not possible either. Engineering success by approximation is therefore to be sought within the domains where precise learning is possible, as in the natural language case. On this basis we discuss how modern language models succeed without the i.i.d.\ condition. They work where precise learning is possible, and massive data lets them approximate well without that condition. We offer this as a clue toward why scaling laws hold \citep{kaplan2020, hoffmann2022}, not as a proof that they do.

Throughout this paper, we restrict the empirical targets to true probabilities of data-generating processes and their values to computable reals. The first restriction matches current practice. A large language model (LLM) that predicts the next word computes an estimate of a true conditional probability, an image classifier outputs probabilities over classes, and the standard textbooks treat learning as the computation of probabilities from the start \citep{Vaswani:17, He:16, Bishop:06, Murphy:12, Bishop:24}. The second restriction, to the computable reals, matches the discrete nature of machine computation, and it keeps the equivalence between algorithmic execution and derivability exact; uncomputable reals such as Chaitin's constant are excluded. A true probability that is not a computable real may exist, but no machine can learn it, and the theory thus separates what exists from what a machine can learn.

\paragraph{Contributions.} The paper makes four contributions.

1. \textit{A new definition of machine learning and a formal system for it} (Section 2)

We define learning as successful computation, and successful computation as the existence of a proper derivation in a formal system $\mathcal{S}_t$ built from the axioms of arithmetic, the mathematical definitions, and the empirical definitions that record the data (Definitions~\ref{def:learn} to~\ref{def:success}). G\"{o}del's Representability Theorem in classical form links computability with provability, and this paper extends it in two ways. First, the domain of the theorem is extended from the natural numbers to the computable reals, through index tracking, so that the theorem covers real-valued functions (Theorem~\ref{thm:rep}). Second, the notion of computability so extended is linked to a formal system that also handles empirical entities, and the classical notion of computability thereby becomes the new notion of successful computability (Definition~\ref{def:success}). 

In short, where G\"{o}del's theorem links computability with provability, our Theorem~\ref{thm:rep} and Definition~\ref{def:success} link successful computability with provability in the extended formal system that has empirical definitions. So the question we ask differs from the one asked in the recent literature on computability and learning \citep{Agarwal:20, Ackerman:22, Crook:22}. That literature takes the PAC model, with its i.i.d.\ assumption, as given, and asks whether a learner that meets the PAC guarantee can be a computable one. Our question comes one step earlier: we ask whether the guarantee itself can be derived from the record without a global assumption, that is, whether learning is carried out inside an extended formal system in which the machine has a proper derivation, from the record, of what it has learned. When the target is the true probability, Section~\ref{sec:pac} shows that a derivation from the i.i.d.\ assumption is circular. So a computable learner in the sense of the recent literature has still not learned the true probability in our sense.

2. \textit{The conditions for learning a true probability, and the circularity of the alternative} (Section 3 and Section 4)

A machine learns a true probability if and only if the Truth Condition and the Assurance Condition are met (Theorem~\ref{thm:cond}). The Assurance Condition cannot be weakened: a machine that assigns its convergence a probability below~1 tolerates the possibility of error infinitely often (Theorem~\ref{thm:tolerate}). A machine whose outputs may be wrong infinitely often cannot be said to learn. The Truth Condition states statistical consistency without a global assumption. Statistical learning theory, by contrast, obtains convergence from a global assumption on the data, i.i.d.\ in PAC learning and Markov or mixing conditions in its relaxations. Such an assumption is not derivable from any finite record, so it can enter $\mathcal{S}_t$ only as an axiom; a derivation from it is a mere derivation, and when the target is the true probability itself it is circular, so that it is not genuine learning (Theorem~\ref{thm:circ}).

3. \textit{A learnable domain, and what it means for real machines} (Section 5 and Section 3) 

The conditions of Theorem~\ref{thm:cond} are met at a finite time for the true probability of a sentence over a finite ideal corpus of natural language (Theorem~\ref{thm:corpus}). A real machine, which approximates rather than learns precisely, can approximate a true probability only if an ideal machine can learn it precisely (Theorem~\ref{thm:approx}). So the learnability of a domain by an ideal machine is a necessary condition of a real machine's success in it, and it is where we look for the reason that scaling laws hold without the i.i.d.\ condition.

4. \textit{The hybrid AI as a consequence} (Section 6) 

Classical proofs and programs were already derivations in a formal system: a proof is a derivation by definition, and a completed computation is a derivation by G\"{o}del's theorem. However, modern AI proves theorems and writes code differently. An LLM proposes a candidate proof or program from probabilities it has computed over past data, and a deductive verifier then checks the candidate. So the task is done by a statistical method and a deductive method used together, and the statistical half is itself the computation of an empirical function. 

Now, we have reformulated empirical learning as a derivation in the same kind of proof system as traditional mathematical reasoning and coding. So a single deductive mechanism serves for all three in our framework: mathematical reasoning, coding, and empirical inference. For the third task in particular, the derivation is deductive but what is derived remains probabilistic, since an empirical target requires the record for derivation. So the extended formal system we propose is a \textit{hybrid} one, in the sense that purely mathematical functions and empirical probabilities are derived from two different kinds of premises, the mathematical and the empirical definitions, although they are all derived by one and the same deductive mechanism.

\section{Machine Learning as Successful Computation}
\label{sec:theory}

This section defines the notion of machine learning in three steps, and it helps to say at the outset which of the steps are definitions and which is a theorem. The empirical targets of this paper are true probabilities of data-generating processes, defined in Section~\ref{sec:true}; the definitions below are stated for empirical functions in general, but the claims of the paper concern these targets.

First, a machine \textit{learns} a function when it \textit{succeeds} in \textit{computing} it (Definition~\ref{def:learn}). Learning is defined through computation because what a machine does is compute. The word ``succeed'' is important here: not every computation that returns the right value counts as learning. What makes computation count is fixed in the third step.

Second, what a machine computes has a \textit{derivation}. By G\"odel's Representability Theorem, a machine computes $f(a) = b$ exactly when an arithmetical formula $\varphi_f(\bar a, \bar b)$ is derivable in a formal system of arithmetic. We will write the numeral of a number with a bar throughout, so $\bar a$ and $\bar b$ here are the numerals that denote the numbers $a$ and $b$ in the formal system. The formula $\varphi_f(\bar a, \bar b)$ is then the sentence that says of the numerals $\bar a$ and $\bar b$ what the equation $f(a) = b$ says of the numbers $a$ and $b$, namely that $b$ is the value of $f$ at $a$. We use G\"odel's theorem in two ways: as the link between computation and derivation, and as the basis of Theorem~\ref{thm:rep}, an extension of the theorem to the computable reals, which can thus cover targets such as true probabilities whose values are real numbers.

Third, a machine succeeds in computing a function when each of its values has a \textit{proper} derivation in a formal system extended by the record of the data, admitted as empirical definitions (Definitions~\ref{def:proper} and~\ref{def:success}). A formal system of arithmetic as ordinarily given has no such definitions for what has been observed, and our extension supplies them. Definition~\ref{def:success} uses Theorem~\ref{thm:rep} twice, for the derivation it asks for and for the real values its targets take, and that is why the theorem is stated first.

We begin with some key terms used by the three steps. By a \textit{machine} we mean a Turing machine, and by \textit{computation} we mean what a Turing machine does. A Turing machine is a finite list of instructions, so every Turing machine can be written as a finite string and hence can be coded by a natural number $e$, such as its G\"odel number. We will call $e$ the \emph{index} of the machine. The numeral $\bar{e}$ inside a formal system then denotes the index $e$. A real number $x$ is computable when some machine, given $n$, prints a rational $q(n)$ with
$$ |q(n) - x| \le \pw{n}, $$
the rate we require throughout. The index of such a machine is also an index of $x$, and we will write it as $e_x$. Note that no real number is ever on the tape of the machine. What is on the tape is an index, and the machine prints an approximation when one is asked for. When a machine computes a function $f$ on reals, what it computes in fact is a function on indices, from the indices of the arguments to an index of the value. We will call this function the \emph{index tracking function} of $f$ and write it as $h_f$; its domain is the set of valid indices of the arguments of $f$.

In classical proving and coding, doing the task amounts to deriving each output of a certain mathematical function in a formal system. (How modern AI systems prove and code by a \textit{statistical proposer} together with a \textit{deductive verifier} is briefly discussed in Section~\ref{sec:concl}. Here we state only the classical account.) A machine has proved a sentence $\sigma$ when it has produced a derivation of $\sigma$ in a formal system. Whether a given string $p$ is a proof of $\sigma$ is decided by a primitive recursive function $\Prf(p,\sigma)$, equal to $1$ if $p$ proves $\sigma$ and to $0$ otherwise, and by G\"{o}del's Representability Theorem the system derives the value of $\Prf$ at $(p,\sigma)$, whichever it is. So checking a proof is deriving an output of a mathematical function. Coding has the same shape. 
Whether the program with index $e$, run on the input $u$, returns the output $v$ within $s$ steps is decided by a primitive recursive function $\Test(e, u, v, s)$, equal to $1$ if it does and to $0$ otherwise, and the system derives its value in the same way. Running a program is the special case in which the program is given rather than sought.

In both tasks, then, what the machine has done is to derive each output of a function. Now, \textit{derivability} is an epistemic notion. \citet{Godel:46} counted computability and provability among the epistemological notions and remarked that Turing's analysis had given the first of them an absolute definition, and \citet{Fefer:06} read provability as what can be known in mathematics. Both writers use ``provable'' and ``knowable'' almost interchangeably: what is derived is what is known. On this reading, a machine knows a function when it derives each output of the function. Learning, in turn, is the acquisition of knowledge \citep{Valiant:84}. So a machine learns a function when it comes to know it, that is, when it properly derives each of its outputs. The epistemic reading of provability was, in G\"{o}del and Feferman, a reading about pure mathematics; we apply it to empirical functions as well in this paper.

Let us now define when machines can be said to learn.

\begin{definition}[Machine Learning]
\label{def:learn}
Machines \textbf{learn} a target function $f^*$ when they succeed in computing $f^*$ if $f^*$ exists.
\end{definition}

The clause ``if $f^*$ exists'' is there for empirical targets. Whether a true probability really exists is a fact about the world, and Definition~\ref{def:learn} does not presuppose it. The existence of a true probability is one thing and its learning, that is, its derivation from the record, is another; the definition speaks of the second without presupposing the first. 

Definition~\ref{def:learn} involves the notion of \textit{success in computation}, which indicates that learnability requires more than computability in the traditional sense. To define this notion we need one thing about computation, namely that whatever a machine computes has a derivation in a formal system of arithmetic (Theorem~\ref{thm:rep}). Given this, Definition~\ref{def:success} says when a computation, in particular the computation of an empirical function, is successful, in terms of the derivation it has in a formal system extended by empirical definitions; this definition in turn depends on the notion of a proper derivation in Definition~\ref{def:proper}. 

So Definition~\ref{def:learn} is based on Theorem~\ref{thm:rep}, Definition~\ref{def:proper} and Definition~\ref{def:success}. We nonetheless state it first, because it is the central definition of our theory and is therefore better made explicit from the start.

We now turn to Theorem~\ref{thm:rep}.

\begin{theorem}[Instance-wise Representability of Computable Reals via Index Tracking; adapted from \citet{Godel:31}]
\label{thm:rep}
Let $\mathbb{R}_c$ be the set of computable real numbers, and let $T$ be a formal system of arithmetic, in the sense explained below. For any computable function $f: \mathbb{R}_c^n \to \mathbb{R}_c$, its index tracking function $h_f$ is defined on the valid indices of the arguments of $f$ and is computable: it is the restriction to those indices of a partial recursive function on $\mathbb{N}^n$, and it halts on each of them.

For any input instance $a = (a_1, \dots, a_n) \in \mathbb{R}_c^n$ and its output instance $b = f(a) \in \mathbb{R}_c$, let $e_a = (e_{a_1}, \dots, e_{a_n}) \in \mathbb{N}^n$ be the specific vector of indices provided to the machine, and let $\overline{e_a}$ be the tuple of their numerals. If the machine's run halts with the output index $e_b \in \mathbb{N}$ (i.e., $h_f(e_a) = e_b$), then $e_b$ is an index of $f(a)$, that is, of $b$, and there exists an arithmetical formula $\varphi_f(x_1, \dots, x_n, y)$ of $T$ such that
$$ h_f(e_a) = e_b \iff T \vdash \varphi_f(\overline{e_a}, \overline{e_b}). $$
\end{theorem}

\noindent In words: whenever the machine's run returns an index of the value when given the indices of the arguments, the system derives the arithmetical statement that it did so, and conversely. The letters $a$ and $b$ denote an instance, one particular input and its output; the variables $x_1, \dots, x_n, y$ of the formula range over numerals; and in Definition~\ref{def:success} below, $x$ and $y$ mean an arbitrary input and output. The theorem is stated instance by instance, and that is the form in which it is used throughout.

Four things are involved here: a formal system of arithmetic, representability, index tracking, and the notation for the result.

\emph{Formal system of arithmetic.} A \textit{formal system} is given by a language and a set of axioms and definitions, together with the inference rules of the underlying logic. A \textit{derivation} in a formal system is a finite sequence of formulas each of which is an axiom, a definition, or what follows from earlier ones by an inference rule, and a \textit{proof} is a derivation whose last formula is the theorem. Whether a given string is a derivation is decided by a mechanical check. A \textit{formal system of arithmetic} $T$ is a formal system that extends Robinson arithmetic $Q$ in the sense that every theorem of $Q$ is a theorem of it; its language therefore contains the language of arithmetic, and it may contain further symbols, as the system $\mathcal{S}_t$ below does. We require in addition that such a system $T$ be consistent and have a recursive set of axioms and definitions. Together with the extension of $Q$, these are the hypotheses of G\"odel's Representability Theorem, and nothing more is needed for Theorem~\ref{thm:rep}.

\emph{Representability.} G\"odel's original theorem concerns functions on the natural numbers. For every recursive $h: \mathbb{N} \to \mathbb{N}$, there is a formula $\varphi_h(x, y)$ of arithmetic such that, for all natural numbers $u$ and $v$ (we use new letters here, since $a$ and $b$ in Theorem~\ref{thm:rep} are reals),
$$ h(u) = v \iff T \vdash \varphi_h(\bar u, \bar v) $$
in any consistent recursively axiomatized $T$ extending $Q$ \citep{Godel:31, Boolos:02}. The formula says, inside the formal system $T$, that the function $h$ sends $u$ to $v$, and the theorem says that $T$ settles each instance exactly as the function $h$ does.

\emph{Index tracking.} A true probability can be a real number, and Theorem~\ref{thm:rep} does not apply to it directly; it applies to its index. In order to compute a function $f(x)$ on real numbers, a machine is not given $x$ but instead an index $e_x$, that is, a program that prints rational approximations to $x$ to any precision. From that program it builds a program that prints approximations to $f(x)$, and it outputs the index of the program it built. For example, let $f(x) = x^2$ on $[0,1]$, the interval in which probabilities lie. Given a program $e_x$ printing $q_x(n)$, the program that on request $n$ prints
$$ q_x(n+2)^2 $$
is within $\tpw{n}$ of $x^2$ (since $|q^2 - x^2| = |q - x|\,|q + x|$ and $|q+x| < 3$ here), so it is a program for $y = x^2$, and writing it down from $e_x$ is a mechanical operation on one number. By the $S^n_m$ theorem \citep{kleene1952introduction}, such an operation is a computable operation on indices, the restriction to the valid indices of a partial recursive function. This is the index tracking function $h_f$ of $f$ introduced above: it tracks, on the natural numbers, what the computation of $f$ does on the reals. Since $h_f$ is a function on natural numbers, representability does apply to it, and that is Theorem~\ref{thm:rep}. So the notion of representability here is G\"odel's, unmodified, and the extension to $\mathbb{R}_c$ is obtained by index tracking, $h_f(e_x) = e_y$.

\emph{Notation.} The computation thus has two layers: the reals are handled through their indices, so that what is computed and what is derived both concern indices. $T \vdash \varphi_{f}(\overline{e_{a}}, \overline{e_{b}})$ means that $T$ derives the formula $\varphi_{f}(\overline{e_{a}}, \overline{e_{b}})$, where $\varphi_{f}$ represents the index tracking function $h_f$ of $f$, and $\overline{e_a}$, $\overline{e_b}$ are the numerals of the indices of $a$ and $b$. So the formula says of the indices what the equation $f(a) = b$ says of the reals: given the index of $a$, the computation of $f$ returned the index of $b$. No real number enters the formula as such. A formula is a finite string of signs, so an infinite object such as a real number can occur in it only through a finite name; symbols such as $\sqrt{2}$ and $\pi$ are names of this kind. Here the name is the numeral $\overline{e_a}$ of an index, which denotes the real through a program that computes it.

\begin{remark}
\label{rem:indices}
Infinitely many Turing machines compute the same real, so an index is not unique. Because $f$ is computable, some machine computes it, and possibly many do. The derivation concerns the machine that was actually run and the path $\overline{e_a} \mapsto \overline{e_b}$ it actually took, so the non-uniqueness does not enter the statement of Theorem~\ref{thm:rep}. See Remark~\ref{rem:B1} in Appendix~\ref{app:remarks}.
\end{remark}

The system $T$ of Theorem~\ref{thm:rep} is a formal system of the classical kind, and in such a system only purely mathematical functions are derived. We now extend it to a system in which empirical functions are derived as well; the extended system is written $\mathcal{S}_t$, and $T$ in Theorem~\ref{thm:rep} may be any formal system of arithmetic, $\mathcal{S}_t$ included.

\paragraph{The system $\mathcal{S}_t$.} In a formal system of arithmetic as ordinarily given, such as $Q$ or Peano arithmetic, every symbol is mathematical and no sentence records an observation. To make the values of an empirical function derivable, we add a third component, namely a set of definitions that introduce symbols for the observables and record their observed values. These definitions are sentences about numbers like any other, since states and values enter as codes and numerals. But what is new is that their symbols stand for observables in the world and their sentences record what has been observed. The result is still a formal system of arithmetic, but one in which a machine operates for both mathematical and empirical targets. For a purely mathematical target, the added definitions are simply not used, as Definition~\ref{def:success} makes explicit.

Since the record of observed facts changes with time, so does the system, and we write the time as a parameter: at any given time $t$, machines operate within $\mathcal{S}_t$, whose language extends the language of arithmetic over the computable reals $\mathbb{R}_c$,
$$ \mathcal{S}_t := \mathcal{A} \cup \mathcal{D}_{math} \cup \mathcal{D}_{emp}(t), $$
where:
\begin{description}[leftmargin=4em, labelindent=25pt, labelwidth=0pt, labelsep=0.5em, font=\normalfont, topsep=2pt, itemsep=2pt, parsep=0pt]
    \item[$\mathcal{A}$ (Axioms):] A recursive set of axioms extending Robinson arithmetic. Such a set suffices for the empirical targets of this paper, and a stronger $\mathcal{A}$ may be taken when the target requires it. The set $\mathcal{A}$ is a parameter of the framework.
    \item[$\mathcal{D}_{math}$ (Mathematical Definitions):] A recursive set of formal definitions for mathematical objects and measures (e.g., distance function or metric).
    \item[$\mathcal{D}_{emp}(t)$ (Empirical Definitions):] A recursive set of formal definitions for empirical objects at time $t$, as follows.
\end{description}
\noindent In order to bridge the semantic reality and the syntactic formal system, $\mathcal{D}_{emp}(t)$ is bipartite:
$$ \mathcal{D}_{emp}(t) := \mathcal{D}_{schema} \cup \mathcal{D}_{fact}(t). $$
\begin{description}[leftmargin=4em, labelindent=25pt, labelwidth=0pt, labelsep=0.5em, font=\normalfont, topsep=2pt, itemsep=2pt, parsep=0pt]
    \item[$\mathcal{D}_{schema}$:] A recursive set of semantic grounding schemas that define how random variables map specific real-world states into syntactic numerals, thereby establishing the well-defined domain over which the true probability measure $P$ is computed.
    \item[$\mathcal{D}_{fact}(t)$:] A finite set of atomic sentences that embed the data filtration $\mathcal{D}_t$ in the syntax, one sentence for each data value physically realized and recorded on the machine's tape up to time $t$. The filtration $\mathcal{D}_t$ is introduced with the ontological filtration $\mathcal{B}_t$ in Section~\ref{sec:true}.
\end{description}
\noindent Since each of these sets is recursive or finite, $\mathcal{S}_t$ is recursive at each $t$, so it is a formal system of arithmetic in the sense above, and Theorem~\ref{thm:rep} applies to it.

With the formal system $\mathcal{S}_t$ in hand, we can now say which derivations in it count. This is the notion of a proper derivation on which Definition~\ref{def:success}, and through it Definition~\ref{def:learn}, depends.

\begin{definition}[Proper and Mere Derivation]
\label{def:proper}
A derivation is a \textbf{proper} one if each of its premises is an axiom in $\mathcal{A}$, a definition in $\mathcal{D}_{math}$ or $\mathcal{D}_{emp}(t)$, or a formula derivable from these. It is a \textbf{mere} derivation if it is valid but one of its premises is a global assumption about the data that is not derivable from them, admitted as though it were an axiom.
\end{definition}

The sentences of $\mathcal{A} \cup \mathcal{D}_{math}$ speak of mathematical objects only, and a purely mathematical target is stated in their language. An empirical target is not: it is defined over the recorded states and values, whose symbols are introduced by $\mathcal{D}_{emp}(t)$, so its formula $\varphi_{f^*}(\overline{e_x}, \overline{e_y})$ contains these symbols, as does any sentence about the world. Now, a sentence that contains these symbols but is still derived without $\mathcal{D}_{emp}(t)$ is a sentence that holds no matter what was observed. For example, it is derivable from logic alone that it will either rain tomorrow or not, and it holds whatever the record says; a derivation of this kind says nothing about the data. The empirical clause of the following definition requires that the value of an empirical target be derived with the help of $\mathcal{D}_{emp}(t)$, and not in this empty way.

\begin{definition}[Successful Computation]
\label{def:success}
Machines successfully compute a target function $f^*$ if and only if, for each input $x \in \mathbb{R}_c$, the output value $y \in \mathbb{R}_c$ generated by them has a proper derivation in $\mathcal{S}_t$:
$$ \mathcal{S}_t \vdash \varphi_{f^*}(\overline{e_x}, \overline{e_y}), $$
where $\varphi_{f^*}$ is the formula representing the index tracking function $h_{f^*}$ of $f^*$, and $\overline{e_x}$, $\overline{e_y}$ are the numerals of the index given to the machine and of the index it returns. In particular, when the target function $f^*$ is empirical, machines successfully compute $f^*$ if and only if the derivation requires $\mathcal{D}_{emp}(t)$. That is, for each input $x \in \mathbb{R}_c$,
$$ \mathcal{A} \cup \mathcal{D}_{math} \nvdash \varphi_{f^*}(\overline{e_x}, \overline{e_y}) \quad \text{but} \quad \mathcal{A} \cup \mathcal{D}_{math} \cup \mathcal{D}_{emp}(t) \vdash \varphi_{f^*}(\overline{e_x}, \overline{e_y}). $$
On the other hand, if $f^*$ is purely mathematical (e.g., prime factorization or calculating the digits of $\pi$), it is successfully computed without needing $\mathcal{D}_{emp}(t)$, i.e., $\mathcal{A} \cup \mathcal{D}_{math} \vdash \varphi_{f^*}(\overline{e_x}, \overline{e_y})$.
\end{definition}

Definition~\ref{def:success} asks for more than Theorem~\ref{thm:rep} gives, and the difference is the point of the definition. Let $g$ be the function on $\mathbb{R}_c$ that the program run by the machine computes, and $h_g$ its index tracking function. By Theorem~\ref{thm:rep}, whenever the program returns $e_y$ given $e_x$,
$$ \mathcal{A} \vdash \varphi_g(\overline{e_x}, \overline{e_y}), $$
which says that the program, given $e_x$, returned $e_y$. This is a statement about the program, and it is derivable from $\mathcal{A}$ alone. Definition~\ref{def:success} asks for a statement about the target,
$$ \mathcal{S}_t \vdash \varphi_{f^*}(\overline{e_x}, \overline{e_y}), $$
which says that $y$ is the value of the target $f^*$ at $x$. The first derivation does not yield the second. Whether the second is available, and from which premises, depends on what determines the values of $f^*$. The values of a purely mathematical target are determined by its definition in $\mathcal{D}_{math}$ (e.g., the digits of $\pi$ are defined there), so the second derivation needs only $\mathcal{A} \cup \mathcal{D}_{math}$, just as the first needs only $\mathcal{A}$. By contrast, the values of an empirical target are determined by what the world has done, which enters $\mathcal{S}_t$ through $\mathcal{D}_{emp}(t)$. For example, the true probability of rain is defined over the recorded states and values (Definition~\ref{def:trueprob} below), and its value is not derivable from $\mathcal{A} \cup \mathcal{D}_{math}$. So the second derivation of an empirical target needs $\mathcal{D}_{emp}(t)$ as well, which is what the empirical clause of Definition~\ref{def:success} requires. 

Note that an empirical function is a function in the mathematical sense like any other, here a measurable function on the probability space of Section~\ref{sec:true}. So the adjective ``empirical'' here concerns what fixes its values, namely the world, and what is derived about such a function remains probabilistic, since its values are probabilities. This difference between the two kinds of target, in the premises necessary for the second derivation, is what the rest of the paper is about.

\paragraph{The empirical definitions in detail.} To illustrate the crucial role of $\mathcal{D}_{schema}$, consider the true probability $P(A_{t+1}|\mathcal{B}_t)$ that it will rain tomorrow. Let the state of our world be either rainy ($\omega_0$) or sunny ($\omega_1$). The event of raining at $t+1$ is defined as the set $A_{t+1} = \{\omega_0\}$. A random variable $X_{t+1}$ is a mathematical function that maps these states to numerals: $X_{t+1}(\omega_0) = 1$ if it rains, and $X_{t+1}(\omega_1) = 0$ otherwise.

In general probability theory, a random variable is merely a measurable function mapping states to numbers. But mathematics alone does not specify which real-world fact (e.g., rain or a coin toss) is represented by the variable. Therefore, we must explicitly define that $X_{t+1}$ represents the fact of raining. This semantic grounding is precisely what $\mathcal{D}_{schema}$ provides. Furthermore, the definition of $X_{t+1}$ fixes the domain of the true probability function. The true probability $P(A_{t+1}|\mathcal{B}_t)$ is formally defined over the pre-image $X_{t+1}^{-1}(1) = \{\omega_i \in \Omega_{t+1} : X_{t+1}(\omega_i) = 1\} = A_{t+1} \in \mathcal{B}_{t+1}$. Thus, the empirical probability function $P(X_{t+1} = 1|\mathcal{B}_t)$ is defined as the true probability of rain.

Inside $\mathcal{S}_t$, the grounding takes the following form. With $\bar{0}$ the code of the rainy state and $\bar{1}$ the code of the sunny state, for example, $\mathcal{D}_{schema}$ contains
$$ \forall x\, \forall y\, \bigl( \mathsf{X}_{t+1}(x) = y \leftrightarrow \delta(x, y) \bigr), \qquad \delta(x, y) := (x = \bar{0} \wedge y = \bar{1}) \vee (x \neq \bar{0} \wedge y = \bar{0}), $$
which reads: the symbol $\mathsf{X}_{t+1}$ sends the code of a state to the value of the variable at that state, and $\delta$ sends the code of the rainy state to $1$ and every other code to $0$. The second clause is there so that $\delta$ assigns a value to every code; without it, a code other than $\bar0$ and $\bar1$ would have no value, and the definition would make $\mathcal{S}_t$ inconsistent, because a function symbol denotes a value at every argument in first-order logic, so the system would derive both $\delta(\bar c, \mathsf{X}_{t+1}(\bar c))$ and its negation for such a code $\bar c$. 

On the other hand, it is observed values that enter $\mathcal{D}_{fact}(t)$. When it rained today and this was written to the tape, for example, $\mathcal{D}_{fact}(t)$ contains the atomic sentence
$$ d_t = \bar{1}, $$
where $d_t$ is a new name that denotes today's recorded value. The target event $A_{t+1}$ itself is never on the tape at prediction time, since the future is not recorded. 

To sum up, $\mathcal{D}_{schema}$ says what the numeral would mean, and $\mathcal{D}_{fact}(t)$ says that it was observed. A derivation of an empirical target draws on both. In general, for each observable $X^{(j)}$, each time $s$, and each recorded value,
\begin{align*}
&\forall x\, \forall y\, \bigl( \mathsf{X}^{(j)}_s(x) = y \leftrightarrow \delta_j(x, y) \bigr) \in \mathcal{D}_{schema}, & &\delta_j \text{ primitive recursive},\ \ \mathcal{A} \vdash \forall x\, \exists!\, y\, \delta_j(x, y), \\
&d^{(j)}_k = \bar{n} \in \mathcal{D}_{fact}(t), & &d^{(j)}_k \text{ a new name, one per record},
\end{align*}
where $\mathsf{X}^{(j)}_s$ is a function symbol new to the language. The first line says that $\mathsf{X}^{(j)}_s$ is introduced by a mechanically checkable condition that assigns exactly one value to each state, so that it defines a function; the second says that each record is an atomic sentence about a newly recorded value denoted by a new name $d^{(j)}_k$. Since the map $j \mapsto \delta_j$ is computable, $\mathcal{D}_{schema}$ is a recursive set. Both lines are definitional extensions of $\mathcal{A} \cup \mathcal{D}_{math}$, so for any formula $\varphi$ in the language of $\mathcal{A} \cup \mathcal{D}_{math}$,
$$ \mathcal{S}_t \vdash \varphi \implies \mathcal{A} \cup \mathcal{D}_{math} \vdash \varphi : $$
the empirical definitions add a vocabulary for the world and a list of what has been recorded, and no new theorem about numbers.

One point about Definition~\ref{def:success} remains. Its condition is a condition on derivations, and whether a given string is a derivation in $\mathcal{S}_t$ is decided by a mechanical check. But whether what $\mathcal{S}_t$ derives is true of the world is a different question. Even when the value of an empirical target has been derived with the help of $\mathcal{D}_{emp}(t)$, the derivation does not by itself settle whether that value is the true probability. This question is addressed by the Truth Condition of Section~\ref{sec:true}.

\section{Machine Learning on the True Probability}
\label{sec:true}

Section~\ref{sec:theory} left one question open: whether a derived value is the true probability. This section answers it. It states the two conditions under which machines learn a true probability (Theorem~\ref{thm:cond}) and then discusses each: why the Truth Condition requires the precise value (Section~\ref{sec:truth}), and why the Assurance Condition requires exactly $1$ (Section~\ref{sec:assurance}).

\begin{definition}[True Probability]
\label{def:trueprob}
The true probability $P(A_{t+1}|\mathcal{B}_t)$ is defined as the conditional distribution of the random variable $X_{t+1}$ given the ontological filtration $\mathcal{B}_t$ generated by $\{X_0, \dots, X_t\}$ on the probability space $(\Omega, \mathcal{F}, p)$. Together with the initial distribution on $\Omega_0$, the joint measure $p$ is uniquely determined by the collection of these conditional probabilities from the stochastic data-generating process. By the data-generating process we mean the probability space $(\Omega, \mathcal{F}, p)$ together with the observables defined below; the true probabilities are the conditional probabilities of $p$, and conversely they determine $p$ together with the initial distribution.
\end{definition}

Definition~\ref{def:trueprob} describes the world and not the system. The world is set out as follows. For each time $t \in \mathbb{N}$ there is an enumerable set $\Omega_t = \{\omega_0(t), \omega_1(t), \dots\}$ of states, indexed by $i$, and
\[
\Omega = \prod_{t} \Omega_t, \qquad \omega(t) \in \Omega_t, \qquad X^{(j)}_t : \Omega_t \to \mathbb{R}_c \ \ (j \in J), \qquad \mathcal{B}_t = \sigma\bigl(X^{(j)}_s : s \le t,\ j \in J\bigr),
\]
\[
A_{t+1} = \{\omega \in \Omega : X_{t+1}(\omega(t+1)) = x\} \in \mathcal{B}_{t+1}, \qquad P(A_{t+1}|\mathcal{B}_t).
\]
In words: a representative path of the world is a sequence $\omega$ whose coordinate at time $t$ is the state $\omega(t) = \omega_i(t)$ for some $i$, and we drop the time when there is no confusion; the observables $X^{(j)}_t$ are random variables, $X^{(j)}_t$ representing the $j$-th observable at time $t$; $\mathcal{B}_t$ is the $\sigma$-algebra generated by the observables up to time $t$; and $A_{t+1}$ is the event that the observable $X_{t+1}$ takes the value $x$ at time $t+1$, the target being its true probability given $\mathcal{B}_t$. The family of observables is taken to separate the states:
\[
\text{for all } \omega_i, \omega_{i'} \in \Omega_t \text{ with } \omega_i \neq \omega_{i'} \text{ there is } j \in J \text{ with } X^{(j)}_t(\omega_i) \neq X^{(j)}_t(\omega_{i'}).
\]
A set of states $C \subseteq \Omega_{t+1}$ is identified with the set of paths $\{\omega \in \Omega : \omega(t+1) \in C\}$ whose coordinate at $t+1$ lies in $C$; this is why $A_{t+1}$ could be written as a set of states in Section~\ref{sec:theory}. In Definition~\ref{def:trueprob}, $X_s$ in $\{X_0, \dots, X_t\}$ stands for the family $(X^{(j)}_s)_{j \in J}$; the target event concerns one observable of the family, which we write $X_{t+1}$. Nature is to generate each data point $X_{t+1}(\omega_i) = x_i$ with the true probability $P$ so that $X_{t+1}^{-1}(x_i) = \{\omega_i \in \Omega_{t+1} : X_{t+1}(\omega_i) = x_i\} = A_{t+1} \in \mathcal{B}_{t+1}$ and $P(X_{t+1}(\omega_i) = x_i | \mathcal{B}_t) = P(A_{t+1}|\mathcal{B}_t)$. Because the observables separate the states, $\mathcal{B}_t$ is the totality of true facts up to time $t$, and it exists independently of the machine's knowledge.

To clarify the record, we must distinguish the ontological filtration $\mathcal{B}_t$ from the data filtration $\mathcal{D}_t$. $\mathcal{B}_t$ is the $\sigma$-algebra of all true facts that have occurred in the world up to time $t$, which determines the true probability $P(A_{t+1}|\mathcal{B}_t)$. In contrast, $\mathcal{D}_t$ is a sub-$\sigma$-algebra of $\mathcal{B}_t$ ($\mathcal{D}_t \subseteq \mathcal{B}_t$), representing only the objective facts recorded as strings of signs (0s and 1s) on the machine's physical tape. That a given value is recorded is itself a fact about the world, whatever the value is, which is why $\mathcal{D}_t \subseteq \mathcal{B}_t$. The framework does not settle when a record is correct, and it does not need to: the machine computes on the record, and what it computes is compared with the truth below.

\begin{definition}[Machine Computation of Subjective Probability]
\label{def:subj}
A machine computes a subjective probability $\Pi(A_{t+1}|\mathcal{D}_t) = \alpha \in \mathbb{R}_c$ at time $t$ if, given the syntactic encoding of the record $\mathcal{D}_t$ as input, it outputs a valid index $e_\alpha$ of $\alpha$, that is, the index of a Turing machine that prints the rational approximations to $\alpha$ at the rate fixed in Section~\ref{sec:theory}. The map from encoded records to indices that the machine so computes is its index tracking function $h_\Pi$.
\end{definition}

The word ``subjective'' marks only that the probability is the machine's output and need not be the true probability; no further theory of subjective probability is assumed. In \citet{Ramsey:31}, a conditional probability expresses an agent's degree of belief in a proposition given a condition, measured by a bet that is called off if the condition fails. Similarly, the machine's subjective probability $\Pi$ is not a product of omniscient knowledge of the true world $\mathcal{B}_t$. It is the output of a physical device executing the tracking algorithm $h_\Pi$ based on the syntactic strings of signs ($\mathcal{D}_t$) recorded on its tape, the records being treated by the machine as true.

The machine's computation is a conditional output: assuming that the record $\mathcal{D}_{emp}(t)$ is true, the machine computes the index $e_\alpha$ and thereby assigns the probability $\alpha$ to $A_{t+1}$. Ramsey used the betting device to measure a person's degree of belief, which is an inner state and cannot be read off directly \citep{Ramsey:31}. Savage later made explicit what the device is for: a person may have an incentive not to report the belief they sincerely hold, and a well-designed bet makes sincere reporting the best choice \citep{Savage:71}. We need none of this machinery. A machine's probability is the output of a computation, so the machine does not deviate from its computation in returning its output; it simply outputs what it computes. What we take from Ramsey is only the conditional form: the probability is computed under the assumption that the condition holds. Thus, the data filtration $\mathcal{D}_t$ need not contain perfect information; the machine computes even under incomplete data. Whether or not anything outside the record is assumed, the machine's computation does not change, since it computes on the recorded data alone. Since $\mathcal{D}_t \subseteq \mathcal{B}_t$, whatever the machine computes from the record is $\mathcal{B}_t$-measurable, and the unrecorded facts $\mathcal{B}_t \setminus \mathcal{D}_t$ do not enter the computation. Thus we obtain
$$ \Pi(A_{t+1}|\mathcal{B}_t) = \Pi(A_{t+1}|\mathcal{D}_t \cup (\mathcal{B}_t \setminus \mathcal{D}_t)) = \Pi(A_{t+1}|\mathcal{D}_t), $$
so we may write $\Pi(A_{t+1}|\mathcal{B}_t)$ for $\Pi(A_{t+1}|\mathcal{D}_t)$. This is a convention of notation and not a claim that the machine conditions on facts it does not have. The convention lets the learning condition compare $\Pi$ and $P$ on the one filtration $\mathcal{B}_t$, and it reads the machine as accepting data from the same world as Nature. However, in order to genuinely learn the true probability, $\mathcal{D}_t$ must contain sufficient empirical data to compute $P(A_{t+1}|\mathcal{B}_t)$, which is manifested as
$$ \Pi(A_{t+1}|\mathcal{B}_t) = \Pi(A_{t+1}|\mathcal{D}_t) = P(A_{t+1}|\mathcal{B}_t). $$

Note that a machine does not output an infinite decimal. Two kinds of machine appear in Definition~\ref{def:subj}, and they work on different levels. The \textit{learner} reads the record on its tape and writes a natural number, the index $e_t$ of a program. The \textit{program} is the second machine: given a precision request $k$, it prints a rational within $\tpw{k}$ of the probability. So the learner does not print the digits of its probability. It writes a program, and the program computes the probability to any required precision. What this means for the statement of convergence is set out after the theorem.

\begin{theorem}[Conditions for Learning]
\label{thm:cond}
Consider any computable real number $\alpha^* \in [0,1]$ and write $\Pi_t := \Pi(A_{t+1}|\mathcal{B}_t)$. Machines learn the true probability $P(A_{t+1}|\mathcal{B}_{t})$ as $\alpha^*$ (achieving Computational Success) if and only if, fixing $P(A_{t+1}|\mathcal{B}_{t}) = \alpha^*$, the following two conditions are met:
\begin{enumerate}[label={(\arabic*)}, topsep=2pt, itemsep=0pt, parsep=0pt]
    \item \textbf{Truth Condition (Objective Correctness):} The estimate converges to the truth along the sample path:
    $$ \limsup_{t \to \infty} |\Pi_t - P(A_{t+1}|\mathcal{B}_t)| < \epsilon, \quad \forall \epsilon > 0 $$
    \item \textbf{Assurance Condition (Subjective Justification):} The machine must be self-assured of the truth condition whenever it holds. That is, the machine assigns subjective probability 1 to its own convergence:
    $$ \Pi \left( \limsup_{t \to \infty} |\Pi_t - P(A_{t+1}|\mathcal{B}_t)| < \epsilon, \quad \forall \epsilon > 0 \right) = 1 $$
\end{enumerate}
\end{theorem}

\begin{figure}[ht]
\centering
\begin{tikzpicture}[>=Stealth, font=\small, scale=0.82, transform shape]
\def\xa{0} \def\xb{2.5} \def\xc{5.0} \def\xd{6.9} \def\xe{9.2}
\node (v0) at (\xa,2.8) {$\Pi_t$};
\node (v1) at (\xb,2.8) {$\Pi_{t+1}$};
\node (v2) at (\xc,2.8) {$\Pi_{t+2}$};
\node (vd) at (\xd,2.8) {$\cdots$};
\node (va) at (\xe,2.8) {$\alpha^*$};
\draw[->] (vd) -- node[above,font=\footnotesize]{$\Pi_t\to\alpha^*$} (va);
\node (e0) at (\xa,0.9) {$e_t:\ q_t(n)$};
\node (e1) at (\xb,0.9) {$e_{t+1}:\ q_{t+1}(n)$};
\node (e2) at (\xc,0.9) {$e_{t+2}:\ q_{t+2}(n)$};
\node (ed) at (\xd,0.9) {$\cdots$};
\node (ea) at (\xe,0.9) {$e_{\alpha^*}:\ q_{\alpha^*}(n)$};
\draw[->] (e0) -- (v0); \draw[->] (e1) -- (v1); \draw[->] (e2) -- (v2); \draw[->] (ea) -- (va);
\draw[decorate,decoration={brace,mirror,amplitude=5pt}] (-0.9,0.5) -- (10.3,0.5)
  node[midway,below=7pt,font=\footnotesize,align=center]{$\forall k\,\exists N\,\forall t\ge N\,\forall n\ \ \lvert q_t(n)-q_{\alpha^*}(n)\rvert \le \tpw{k}+\tpw{n-1}$\\[2pt] the arithmetical formula \eqref{eq:conv} the machine derives in $\mathcal{S}_t$ (Assurance Condition)};
\node[anchor=east,align=right,font=\footnotesize] at (-1.3,2.8) {values in $\mathbb{R}_c$: what we state\\in the metatheory (Truth Condition)};
\node[anchor=east,align=right,font=\footnotesize] at (-1.3,1.85) {the program with index $e$\\computes the value};
\node[anchor=east,align=right,font=\footnotesize] at (-1.3,0.9) {indices written by the learner from the record;\\rationals printed by the indexed programs};
\end{tikzpicture}
\caption{The two levels. The learner writes indices; the programs denoted by them print rationals; the values computed by them, and their convergence, are stated in the metatheory.}
\label{fig:levels}
\end{figure}
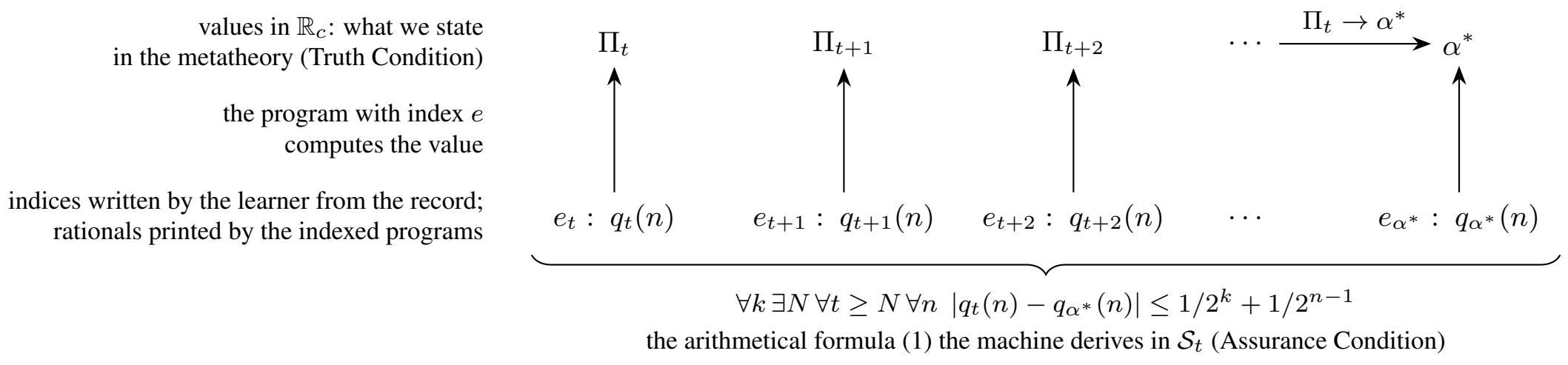

The Assurance Condition asks the machine to derive its own convergence in $\mathcal{S}_t$. But Theorem~\ref{thm:cond} states convergence in terms of the reals $\Pi_t$ and $\alpha^*$, and no real number enters a formula of $\mathcal{S}_t$ (Section~\ref{sec:theory}). So the convergence must be stated as a sentence about indices. What the machine derives is an arithmetical formula whose quantifiers range over natural numbers; no limit enters as an object. Let $e_t$ be the index written by the learner at stage $t$, let $q_t(n)$ be the rational printed by the program with index $e_t$ on the request $n$, and let $e_{\alpha^*}$ and $q_{\alpha^*}(n)$ be the same for a fixed program that computes $\alpha^*$. Convergence of $\Pi_t$ to $\alpha^*$ is then the formula
\begin{equation}
\label{eq:conv}
\forall k\, \exists N\, \forall t \ge N\, \forall n \ \ \bigl| q_t(n) - q_{\alpha^*}(n) \bigr| \le \pw{k} + \pw{n-1},
\end{equation}
in which $q_t(n) = q$ abbreviates $\exists s\,(T(e_t, n, s) \wedge U(s) = \bar{c})$ in Kleene normal form, with $c$ the code of the rational $q$. In words: for every $k$, from some stage on, the value computed by each program written by the learner lies within $\tpw{k}$ of $\alpha^*$. The formula states this through the rationals printed by the programs, and Appendix~\ref{app:remarks} shows that the two statements are equivalent. Figure~\ref{fig:levels} shows the three roles: the learner writes the index, the program computes the value, and the statement that the values converge in $\mathbb{R}_c$ is made by us, the provers of Theorem~\ref{thm:cond}, in the metatheory. So the two conditions of Theorem~\ref{thm:cond} are stated at the upper level, and what the machine derives inside $\mathcal{S}_t$ is formula \eqref{eq:conv}, at the lower level. Remark~\ref{rem:B2} in Appendix~\ref{app:remarks} gives the details.

\subsection{The Truth Condition: precise learning and approximation}
\label{sec:truth}

Let us begin by restating the Truth Condition. Fixing $P(A_{t+1}|\mathcal{B}_t) = \alpha^*$,
$$ \limsup_{t \to \infty} |\Pi_t - \alpha^*| < \epsilon \quad \text{for every } \epsilon > 0 . $$
This condition is the same as \textit{statistical consistency}: the estimate converges to the true probability along the sample path. We state it without any global assumption about the data. It asks for the precise value in the limit, to every precision, and it is fair to ask why. Given that real machines have computational limitations in handling data, it is unlikely that they can compute true probability functions with infinite precision. Instead, real machines in contemporary machine learning aim for \textit{approximations}, which have nevertheless led to remarkable success in language and in vision \citep{Vaswani:17, He:16}. So it is legitimate to ask why our theory requires machines to compute true probabilities \textit{precisely} to learn. Does this imply that our theory is too narrow to cover real-world machine learning practices? To answer this, let us introduce basic notions defining when machines perform approximations.

Let $(\Xi, \mathcal{G})$ be a measurable space, where $\mathcal{G}$ is a $\sigma$-algebra generated by the random vector $X$ in $\Xi$. We refer to $\Xi$ as the \textit{sample space} of $X$ and fix a space $\Psi$, the \textit{value space} of some random variable $Y$. Suppose that there is a true target function $f^{\ast}: \Xi \rightarrow \Psi$ whose learning is the task given to the machine.

\begin{definition}[Learning Rule] \label{def:rule}
A \textbf{learning rule} is any sequence of measurable functions, $f_{n,m}$, provided by machines, where each
\[ f_{n,m}: (\Xi^m \times \Psi^n) \times \Xi^{\infty} \rightarrow \Psi^{\infty} \]
depends on the following data:
\begin{enumerate}[label={(\arabic*)}, topsep=2pt, itemsep=0pt, parsep=0pt]
    \item the initial training samples of labeled data $\{(X_{i},Y_{i})\}_{i=1}^{n}$
    \item the unlabeled data $\{X_{i}\}_{i=n+1}^{m}$
    \item the sequence of future random variables $\{X_{m+k}\}_{k=1}^{\infty}$ for prediction, given that $m \ge n$ for $n,m \in \mathbb{N} \cup \{0\}$.
\end{enumerate}
\end{definition}

The outputs of $f_{n,m}$ are predictions of the future values $\{Y_{m+k}\}_{k=1}^{\infty}$ associated with the future sequence $\{X_{m+k}\}_{k=1}^{\infty}$, derived from the given data. If $m=n$ with $n>0$, the rule corresponds to \emph{supervised learning}; if $m>n$ and $n>0$, it corresponds to \emph{semi-supervised learning}; and if $m>n$ but $n=0$, it corresponds to \emph{unsupervised learning}.

As noted by \citet{Bishop:06} and \citet{Murphy:12}, when the target function is the true probability itself, the task is fundamentally an unsupervised learning problem. In this scenario, Nature does not reveal the exact value of the true probability function at each time step (it only reveals the realized sample). Therefore, learning true probabilities is fundamentally distinct from standard online learning or regret-minimization frameworks, where the true loss or label is revealed sequentially.

\begin{definition}[Distance Function] \label{def:dist}
A \textbf{distance function} is a functional $\ell: \Psi^{\infty} \times \Psi^{\infty} \rightarrow [0, \infty)$ such that:
\begin{equation*}
\ell \bigl( f_{n,m}(\{X_{i}\}_{i=1}^{m}, f^{\ast}(\{X_{i}\}_{i=1}^{n}), \{X_{i}\}_{i=m+1}^{\infty}), f^{\ast}(\{X_{i}\}_{i=1}^{\infty}) \bigr).
\end{equation*}
\end{definition}

In statistical learning theory, $\ell$ is often called a \emph{loss function}, connected to the notion of risk when data come from \textit{i.i.d.} sources. Here, we do not require $\ell$ to be symmetric, as some measures of ``distance'' between probability functions (e.g., Kullback--Leibler divergence) are not a metric. Some other well-known measures of distance between probability functions include Variational Distance, Brier Score, and Hellinger Distance. Formally, $\ell$ captures the distance between the predicted sequence generated by the learning rule $f_{n,m}$ and the true sequence generated by the target function $f^{\ast}$.

\begin{definition}[Approximation] \label{def:approx}
Let $\delta > 0$ be a given error bound. Machines \textbf{approximate} the target function $f^{\ast}$ within $\delta$ by a learning rule $f_{n,m}$ if they successfully compute that $\ell(f_{n,m}(\cdot), f^{\ast}) < \delta$, that is, if there is a proper derivation in $\mathcal{S}_t$ that the rule lies within $\delta$ of the target.
\end{definition}

\begin{theorem} \label{thm:approx}
Suppose that no \emph{ideal} machine can learn the target function $f_{\infty}^{\ast}$, even from the infinite series of training samples $\{X_{i}\}_{i=1}^{\infty}$. Then there is an error bound $\delta_0 > 0$ such that, for every $\delta < \delta_0$ and every finite learning rule $f_{n,m}$ (with $n,m<\infty$), no \emph{real} machine can successfully compute that $f_{n,m}$ lies within $\delta$ of $f_{\infty}^{\ast}$. Hence no real machine approximates $f_{\infty}^{\ast}$ within any $\delta < \delta_0$.
\end{theorem}

The bound $\delta_0$ is needed because a large $\delta$ can be met trivially. A probability lies in $[0,1]$, so a learning rule that always outputs $1/2$ lies within any $\delta > 1/2$ of every true probability, and this can be derived without any data. An ideal machine here is a classical Turing machine with no bound on time and no bound on tape. It is not a machine of any other kind, and in particular it is not a machine with access to uncomputable objects. 

The idea behind Theorem~\ref{thm:approx} is the one that separates, for example, the computable real $\pi$ from a non-computable real number. Because $\pi$ is computable, a real machine given any $\delta$, however small, can produce a rational and establish that the rational lies within $\delta$ of $\pi$. For a non-computable real, however, there is a bound $\delta_0$ below which no machine can do this, whether it is an ideal machine or a real one. It may output rationals, and some of them may happen to be close, but no machine can establish that the distance between any of them and the number is below $\delta$. Theorem~\ref{thm:approx} says that a target which is not learnable by any ideal machine is like such a number: a real machine may come close to the number, but no real machine can establish that it has. Probabilistic bounds, which say that a learning rule lies within $\delta$ with some probability, are the subject of Section~\ref{sec:pac}, where we show from what premises such a bound can be derived when the target is the true probability.

Under a global assumption about the data, such as the i.i.d.\ assumption, PAC learning proves a probabilistic bound. Applied to a true probability, the derivation is a mere derivation in the sense of Definition~\ref{def:proper}, and it is circular as well (Theorem~\ref{thm:circ}, Section~\ref{sec:pac}). Without such an assumption, Theorem~\ref{thm:approx} makes precise learning by an ideal machine a necessary condition for approximation by a real machine, for every error bound below $\delta_0$. Theorem~\ref{thm:cond} says when that condition is met, and so it fixes the room for practical approximation. An approximation in polynomial time must meet every small $\delta$, so the condition is the same for it: a real machine can approximate a target within polynomial time \textit{only if} an ideal machine can learn it precisely. Whether this condition is met is a feature of the target, not of the machine, so the question becomes which targets are precisely learnable by an ideal machine. Section~\ref{sec:nlp} gives one, namely the probability of a sentence fixed by a finite ideal corpus, and says what Theorem~\ref{thm:approx} then implies for real machines in that case.

\subsection{The Assurance Condition: why exactly 1}
\label{sec:assurance}

Let us begin by restating the Assurance Condition. Fixing $P(A_{t+1}|\mathcal{B}_t) = \alpha^*$,
$$ \Pi \left( \limsup_{t \to \infty} |\Pi_t - \alpha^*| < \epsilon \ \text{ for every } \epsilon > 0 \right) = 1 \quad \text{whenever the Truth Condition holds.} $$
The core of the Assurance Condition is not merely outputting $\Pi=1$ but the \textit{necessity} of the machine's assurance whenever the Truth Condition holds. `Assurance' here simply means the mechanical necessity of the machine's syntactically proving its own convergence, formula \eqref{eq:conv}, within $\mathcal{S}_t$ based on $\mathcal{D}_{emp}(t)$, $\mathcal{A}$, and $\mathcal{D}_{math}$. If the machine converges to the truth but cannot find the syntactic path ($h_\Pi$) to \textit{properly prove} it within $\mathcal{S}_t$, it cannot output $\Pi=1$ on its own convergence. This is the criterion that mathematically separates successful computation (genuine learning) from mere computation (lucky guessing). What the condition rules out is the machine that converges and cannot prove that it does.

The requirement is $\Pi = 1$ rather than a high threshold such as $0.9$ because it is not a threshold at all. A finite machine cannot verify convergence by inspection, since that would mean checking infinitely many future stages. A derivation is a finite sequence of formulas whose premises come from one fixed formal system, so no single derivation can run through an infinite sequence of ever stronger systems. The extensional route to a non-repeating irrational true probability $\alpha^*$, digit by digit, would need the infinite union $\mathcal{S}_t^{(\infty)}$ of ever stronger systems, and that union is not a formal system in which any machine works; the progressions of theories studied by \citet{turing1939systems} and \citet{feferman1962transfinite} are of this kind. 

So computing $\alpha^*$ by the extensional limit is impossible. Consequently, the only logical possibility for a machine to successfully compute (that is, learn) $\alpha^*$ is to intensionally prove the infinite convergence of its algorithm ($h_\Pi$) within a fixed formal system $\mathcal{S}_t$. The proof of Theorem~\ref{thm:cond} shows that the same holds when $\alpha^*$ is rational. A derivation either exists or it does not, so there is no intermediate value for the condition to take. This necessity of a finite syntactic proof of infinite convergence is why the Assurance Condition ($\Pi=1$) is an indispensable requirement for machine learning. 

If the above argument is right, learning true probabilities needs a new foundation that is not supplied by statistical learning theory. That theory accepts asymptotic convergence, our Truth Condition, under the i.i.d.\ assumption. Our framework asks in addition for the Assurance Condition, a derivation of the convergence. The Truth Condition concerns what the machine computes, and the Assurance Condition concerns what the machine can establish about its own computation. The framework thus requires an \textit{epistemic} condition in addition to a computational one.

The Assurance Condition may look paradoxical, since it assigns probability $1$ to infinite future convergence on the basis of finite data. What looks paradoxical is not that convergence over infinitely many stages is stated in a finite formula, for a formula with a universal quantifier does that. What looks paradoxical is that the universal formula is derived from finite premises about the past, the record, without any assumption about the future. Our answer is that whether such a derivation exists depends on the domain. By Theorem~\ref{thm:cond}, genuine learning occurs exactly where it does, and Section~\ref{sec:nlp} exhibits a domain where it does. Remark~\ref{rem:B2} in Appendix~\ref{app:remarks} sets this out. 

One might also argue that our Assurance Condition ($\Pi=1$) is too restrictive, prohibiting a machine from admitting that it might make errors. However, our Assurance Condition demands certainty regarding \textit{asymptotic convergence}, not immediate perfection. The condition does not preclude a machine from admitting that it might make errors initially, provided that it can establish that its estimates will eventually approximate the true probability to any required precision.

\begin{definition}[Toleration of the Possibility of Error] \label{def:tolerate}
Machines \textbf{tolerate the possibility of error} at time $t$ while pursuing the goal of learning the true probability $P(A_{t+1} \mid \mathcal{B}_{t})$, when $\Pi (A_{t+1} \mid \mathcal{B}_{t}) = \alpha$ but $\Pi (\ P(A_{t+1} \mid \mathcal{B}_{t}) \neq \alpha\ ) > 0 $ for some $\alpha\in[0,1].$
\end{definition}

\begin{theorem} \label{thm:tolerate}
If machines satisfy the Truth Condition (1) but not the Assurance Condition (2) of Theorem~\ref{thm:cond}, then they tolerate the possibility of error infinitely many times with their forecast $\alpha$.
\end{theorem}

Theorem~\ref{thm:tolerate} further clarifies why we strictly require the Assurance Condition (i.e., subjective probability 1) rather than a merely high subjective probability (e.g., 0.9). If $\Pi < 1$, a machine must tolerate the possibility of error infinitely often, even if it happens to be correct in all those cases. If so, the machine remains a lucky guesser rather than a genuine learner.

One point about the derivation that assurance requires belongs here. A derivation that goes through from a global assumption admitted as an axiom does not supply the syntactic path. It goes through whatever the record says, so it does not require $\mathcal{D}_{emp}(t)$, and it is a mere derivation in the sense of Definition~\ref{def:proper}. When the target is the true probability, Section~\ref{sec:pac} shows that it is circular as well. The path that assurance requires is a proper derivation in $\mathcal{S}_t$ that requires $\mathcal{D}_{emp}(t)$.

Because it is not sufficient for learning to satisfy the Truth Condition without the Assurance Condition, we argue that injecting any global \textit{empirical} assumption (such as \textit{i.i.d.}) into $\mathcal{S}_t$ at finite $t$ cannot be syntactically justified. Relying on such unprovable assumptions reduces the process to mere computation. In the next section, we critically examine how the i.i.d. assumption in PAC learning leads to epistemic circularity.

\section{Beyond Statistical Consistency: The Epistemic Circularity of PAC Learning}
\label{sec:pac}

According to Definition~\ref{def:learn}, machines learn a target function when they attain computational success with respect to the function. The key idea of Definition~\ref{def:learn} is that successful computation fundamentally differs from mere computation.

\begin{definition}[Mere Computation]
\label{def:mere}
Machines \textbf{merely compute} an empirical target function $f^*$ when, for each input $x$, the derivation of $\varphi_{f^*}(\overline{e_x}, \overline{e_y})$ that they have is a mere derivation in the sense of Definition~\ref{def:proper}: it proceeds from a global assumption about the data, admitted into the axioms as though it were an arithmetical axiom, and it does not require $\mathcal{D}_{emp}(t)$.
\end{definition}

Definition~\ref{def:mere} is the exact complement of the empirical clause of Definition~\ref{def:success}. A global assumption about the data is not derivable from any finite record, so if it enters $\mathcal{S}_t$ at all, it enters as an axiom. The i.i.d.\ assumption of Theorem~\ref{thm:circ} is the case we have in mind.

The clause ``does not require $\mathcal{D}_{emp}(t)$'' concerns the justification, not the arithmetic of the estimate. A PAC machine reads its sample, and the sample is recorded in $\mathcal{D}_{emp}(t)$. What the record does not do is supply the derivation of the correctness claim. With the global assumption among the axioms, the bound that certifies the output is derivable uniformly in the record: it holds whatever the tape says, because the assumption already contains the empirical content that the record was supposed to supply. A justification that is insensitive to the evidence cannot certify a fact about the world. So when the output happens to match the target, the derivation does not certify the match, and the machine is at best accidentally right. This is why mere computation falls short of successful computation.

\begin{definition}[Epistemic Circularity]
\label{def:circ}
A derivation is \textbf{epistemically circular} when the derivation that a target has been learned requires as a premise a global claim about the target itself, so that what was to be derived is already among the premises.
\end{definition}

In logic this structure is called \textit{begging the question}: a derivation that assumes $A$ and then concludes $A$. We add the word ``epistemic'' because the loop here occurs while a machine learns the target. A circular derivation is a mere derivation, since the claim about the target is a global assumption about the data, and Theorem~\ref{thm:circ} shows when a mere derivation is circular as well.

In statistical learning theory, extensive efforts have been made to derive statistical consistency under various global empirical assumptions, ranging from strict i.i.d.\ to progressively relaxed forms such as Markovian or mixing conditions \citep{vapnik1998, aldous1990, yu1994, vidyasagar2003, hanneke2021, blanchard2022}. We do not doubt that these relaxations represent impressive mathematical achievements. However, from the perspective of our formal framework $\mathcal{S}_t$, any reliance on global assumptions raises a fundamental epistemic question regarding machine learning.

Rather than delving into detailed proof-theoretic analyses for every relaxed condition, we focus on the standard PAC learning system \citep{Valiant:84} as a foundational and representative example. PAC learning is a theory of approximation with bounds, not of precise learning. Its guarantee implies statistical consistency under the i.i.d.\ assumption, since it gives a finite sample complexity for every error and every confidence, and it is where that assumption does its work most visibly. By Theorem~\ref{thm:approx}, a bound on approximation has content only where precise learning is possible; what we show here is that, when the target is the true probability, the error against which the bound is stated cannot be computed without presupposing the target, so neither can the bound. We demonstrate how the computation of true probabilities under the standard i.i.d.\ assumption in the PAC framework falls into epistemic circularity, thereby rendering it a mere computation rather than genuine learning.

\begin{theorem}
\label{thm:circ}
When the target concept function is the true probability itself, relying on the i.i.d.\ assumption within the PAC learning system leads to epistemic circularity.
\end{theorem}

The claim in short is this. The PAC error
$$ \mathrm{err}(g) = \Pr_{z \sim D}\bigl[ g(z) \neq f(z) \bigr] $$
counts how often the machine is wrong. When the target is the true probability, the machine is wrong at a stage when its forecast differs from $P(X_{t+1}|\mathcal{B}_t)$ at that stage. To count such stages, the machine must know at each stage whether its forecast equals the true probability, and that is what it set out to learn. The outcome $X_{t+1}$ does not settle it, since either outcome is compatible with either value of the true probability. The i.i.d.\ assumption supplies what is missing. It says that the samples of the true probability are identically distributed, that is, that they all come from stages at which the true probability has a single value. Sorting the stages by that value is exactly what the machine cannot do before it has learned the value. So the error, and with it any bound on the error, is computable only on a premise about the target. That is what Definition~\ref{def:circ} means by ``epistemic circularity'', and the proof in Appendix~\ref{app:proofs} exhibits the loop.

The derivation still goes through, as a derivation. It is valid, and what the machine does is therefore a computation. But what it does not do is learn, since it presumes what it aims at. There is no guarantee that a PAC machine approaches the true probability correctly, and at best it is \textit{lucky} to be correct. This is why learning must go beyond statistical convergence.

The same applies to the computable version of PAC learning \citep{Agarwal:20, Ackerman:22, Crook:22}, which asks whether a learner that meets the PAC guarantee can be a computable one. The guarantee it starts from is the one derived under the i.i.d.\ assumption, so when the target is the true probability the circularity of Theorem~\ref{thm:circ} is inherited, whatever the computability of the learner. Our question, whether the guarantee can be derived from the record at all, comes before theirs.

It is important to acknowledge the immense practical utility of the PAC learning framework. For real machines, PAC learning is highly valuable because it determines the specific sample complexity required to achieve a given error bound within polynomial time. However, our proof of epistemic circularity operates at the fundamental level of computability, independent of whether the learning is analyzed under polynomial-time constraints or general computability without time limits. The implication of our proof is that while the i.i.d.\ assumption is mathematically convenient for calculating specific complexity bounds, this advantage becomes epistemically unjustified when the objective is to learn the true probability itself.

Theorem~\ref{thm:circ} closes one route to the guarantee, the route through a global assumption admitted as an axiom. It does not say that no route exists. What remains is a question of fact: is there a domain in which the two conditions of Theorem~\ref{thm:cond} are met, so that the derivation draws on the record rather than on an assumption about the record? The next section exhibits one, and says what a real machine, which approximates rather than learns precisely, inherits from such a domain.

\section{Bounding the Learnable Space: A Case Study in NLP}
\label{sec:nlp}

In the proof of Theorem~\ref{thm:cond}, we demonstrated how the necessary and sufficient conditions for learning are satisfied when the true probability $P(A_{t+1} \mid \mathcal{B}_t)$ takes a rational value that can be represented in finite time. In this section, we provide a case study of language learning to illustrate this mechanism.

Deep learning has made impressive advances in various aspects of natural language processing (NLP), including speech recognition \citep{hinton2012}. We propose that much of this progress can be attributed to the fact that the true probability of obtaining a specific sequence of words is, in principle, a learnable target under Definition~\ref{def:learn}. As a straightforward example, consider simple (unsmoothed) $N$-gram models to illustrate how true probabilities can be learned. According to \citet{jurafsky2008}, $N$-gram models are central to tasks in which machines must identify words from noisy or ambiguous inputs. In speech recognition tasks, for instance, machines receive input in the form of ambiguous speech and must recognize the sequence of words within that speech. Whenever such speech has been used in actual communication, there must exist a corresponding true population containing the disambiguated form of the input. We refer to this true population as an \textit{ideal corpus}, defined as follows:

\begin{definition}[Ideal Corpus]
\label{def:corpus}
At any given time $t_0$, let us consider the set $W^*$ of all words and sentences that are employed by any group of communicators in our world (e.g., human users) over some time interval ending at $t_0$. We call this set $W^*$ of all words and word sequences the \textbf{ideal corpus}.
\end{definition}

The corpus in Definition~\ref{def:corpus} is ideal in the sense that it includes all relevant words and sentences that have been circulated within any community in this world for certain periods up to any given time. It goes beyond any simple online collection of words and sentences. (Here, words may also represent various ranges of symbols such as numerals, punctuation marks, fragments, filled pauses, tabs, or spaces).

Now, for any \textit{given} sequence of words $S$, what is the probability of obtaining $S$? This is the question we address here. To maintain theoretical consistency with our formal framework, we denote this true probability as $P(S \mid \mathcal{B}_t)$, emphasizing that the probability of a sentence being uttered is inherently conditioned on the ontological filtration $\mathcal{B}_t$, the totality of historical and linguistic contexts established up to time~$t$.

Note that this question is fundamentally different from the Markov question \citep{markov1913}: What is the probability of obtaining any \textit{arbitrary} sequence of words? The Markov question concerns sequences of unbounded length, which no finite corpus contains. This distinction is crucial because it indicates that the way machines learn to recognize sentences in a simple $N$-gram model differs from the way humans learn natural language. The ideal corpus $W^*$ in the simple $N$-gram model contains only a finite number of sentences, whereas humans are supposedly capable of recognizing a potentially infinite number of sentences. Appendix~\ref{app:remarks} discusses the difference between the two questions further. LLMs process unseen sentences, that is, sentences that never occurred in any corpus. Theorem~\ref{thm:corpus} below does not cover such a sentence. Its probability is not fixed by the record of $W^*$, so no derivation from $\mathcal{D}_{emp}(t)$ alone can learn it. We do not treat this question here. Appendix~\ref{app:remarks} states what is missing as Open Problem~\ref{op:productivity}.

Compared with LLMs, the $N$-gram model may look like a toy example. It is chosen because it isolates what makes a true probability learnable in the sense of Definition~\ref{def:success}. The probability of a sentence is fixed by the record of what has actually been used, so the machine derives it from $\mathcal{D}_{emp}(t)$ and not from an assumption about the record. That is the condition which Theorem~\ref{thm:circ} showed the i.i.d.\ route cannot meet, and here it is met.

\begin{theorem}
\label{thm:corpus}
At any given time $t$, there exists an ideal corpus $W^*$ containing a finite number of words and sequences of words actually employed by human users up to time $t$. For any given sentence $S$, the true probability of obtaining $S$ is computable from this corpus $W^*$ by machines. Furthermore, this true probability $P(S \mid \mathcal{B}_t)$ can be learned by the machines.
\end{theorem}

Theorem~\ref{thm:corpus} uses only the finiteness of $W^*$. Its role is to witness that the learnable space under Theorem~\ref{thm:cond} is not empty, and we record that role as a proposition.

\begin{proposition}
\label{prop:nonvac}
The conditions of Theorem~\ref{thm:cond} are satisfiable. In particular, they are satisfied for the target $P(S \mid \mathcal{B}_t)$ over a finite ideal corpus.
\end{proposition}

Theorem~\ref{thm:corpus} shows that an ideal machine can learn the true probability $P(S \mid \mathcal{B}_t)$ for any given sentence $S$. However, in real-world applications where real machines face various technological and resource constraints, they may not be able to learn the true probability $P(S \mid \mathcal{B}_t)$ precisely. Instead, they approximate $P(S \mid \mathcal{B}_t)$ through methods such as maximum likelihood estimation, smoothing techniques, or various clustering algorithms. Regardless of the specific approximation technique employed, it is theoretically significant to establish that such approximations by real machines can, in principle, perform well only when the true probability is precisely learnable by ideal machines, which is what Theorem~\ref{thm:approx} states.

Theorem~\ref{thm:corpus} concerns an ideal machine with the ideal corpus on its tape. No real machine has that corpus. It has a sample of the corpus, and from the sample it must estimate a probability fixed by the ideal corpus. Statistical learning theory guarantees such an estimate when the sample is drawn i.i.d., and the experiments of \citet{ranga2026} show that a transformer can learn the underlying probabilities of synthetic data when the synthetic sample is made i.i.d.\ by construction, drawn from the generating distribution. A transformer is therefore a machine that approximates the true probability well where the i.i.d.\ condition holds. However, the corpus on which a real transformer is trained is not drawn that way \citep{paruchuri2024, meister2021}. Its success in practice then depends on two things. It works in a domain where precise learning is possible, which by Theorem~\ref{thm:approx} is what makes approximation possible at all. And it approximates well without the i.i.d.\ condition, and processing massive amounts of data is what lets it do so. We call the first the \textit{epistemic allowance} of a domain: how much approximation a domain affords a real machine is fixed by what an ideal machine can learn there precisely. We offer the second as a clue toward understanding why scaling laws hold empirically \citep{kaplan2020, hoffmann2022}, not as a proof that they do.

\section{Conclusion}
\label{sec:concl}

We have proposed a formal system in which proving, coding, and learning from data are carried out by a single deductive mechanism, and we have shown that learning a true probability inside it requires more than statistical consistency. We defined learning as successful computation, and successful computation as the existence of a proper derivation of each value in $\mathcal{S}_t$ (Definitions~\ref{def:learn} and~\ref{def:success}). Given these definitions, we can prove which conditions are necessary and sufficient for learning a true probability. By Theorem~\ref{thm:cond}, these conditions are the Truth Condition and the Assurance Condition. The Truth Condition is statistical consistency, and we stated it without a global assumption. The Assurance Condition is what is newly required by our framework, and it is an epistemic demand: a finite derivation of convergence inside the system. By Theorem~\ref{thm:tolerate}, it cannot be weakened to a probability below~1. Theorem~\ref{thm:circ} shows what happens when statistical consistency is derived from a global assumption instead. The derivation goes through and the machine computes, but it is a mere derivation and circular, so the machine does not learn. That is the sense in which learning, as our title says, lies beyond statistical consistency. Theorem~\ref{thm:corpus} shows that both the Truth Condition and the Assurance Condition are met for the probability of a sentence over a finite ideal corpus, and Theorem~\ref{thm:approx} shows that such a domain is also where the approximations of a real machine have content, which is what is exploited by an LLM trained on massive data.

The \textit{hybrid AI} of our title is a consequence of this result and of the definitions behind it. Section~\ref{sec:theory} showed that proving and coding are derivations of the values of the mathematical functions $\Prf$ and $\Test$ from the axioms and the mathematical definitions. Definition~\ref{def:success} makes learning from data a derivation in the same system, one that requires the empirical definitions as well. So the three tasks are done by one and the same deductive mechanism in a single formal system $\mathcal{S}_t$. Still, they are distinguished by the kinds of premises from which they are derived. These two facts together are what we mean by ``hybrid AI''. Accordingly, a hybrid AI in our sense is \textit{not} a combination of a statistical module and a deductive module placed side by side. It is one mechanism with two different kinds of premises, and the two kinds correspond to two kinds of target. The values of a purely mathematical target are fixed by its definition, so they are derivable from the axioms and the mathematical definitions alone. The values of an empirical target are fixed by the world, while the empirical definitions supply the vocabulary for it and the record of what has been observed; whether the values are derivable from that record depends on the domain.

Our hybrid framework also explains what the deductive part of current practice does and what it cannot do. Consider an LLM that proves a theorem with a proof assistant such as Lean. The system has two parts: a \textit{statistical proposer}, which produces a candidate proof from the probabilities it has computed over a record of past proofs, and a \textit{deductive verifier}, which checks whether the candidate is indeed a proof of the sentence. The verifier derives the value of $\Prf$ at the candidate and the sentence, and this derivation is from the axioms and the mathematical definitions alone. So it is a proper derivation and it settles the question with certainty. On the other hand, the probabilities of the proposer are an empirical function, and current practice asks for no derivation of them. Fortunately, none is needed for proving and coding. The target for these tasks is a mathematical function, and the verifier derives its value whatever the proposer did. For proving, a candidate that is not a proof is conclusively rejected, and one that is a proof is conclusively accepted, by the verifier. So the correctness of the output does not depend on whether the proposer has \textit{learned} its probabilities in our sense. The same applies to coding, with $\Test$ in place of $\Prf$.

However, the case is different when the target is itself an empirical function, such as a true probability. There is no verifier here. Whether a candidate value is the true probability is not the value of any mathematical function, since the true probability is fixed by the world and not by the mathematical definitions (Section~\ref{sec:theory}). Nor does the next observation decide it. What is observed is only a definite outcome, such as \textit{rain} or \textit{no rain}, and not a probability. The same outcome can occur whether the true probability is, say, $0.3$, $0.5$, or $0.9$  (Section~\ref{sec:pac}).

So no check from outside the derivation is available inside $\mathcal{S}_t$, and the machine must derive what it outputs itself, with the help of the record. This is why a derivation that is not asked for by current practice becomes a requirement when the target is empirical. The Assurance Condition is that derivation: it does for an empirical target what the verifier does for a mathematical one. Theorem~\ref{thm:cond} says when it exists, Theorem~\ref{thm:circ} says that a derivation from a global assumption does not supply it, and Theorem~\ref{thm:corpus} exhibits a domain where it exists. Our hybrid framework thus asks for a derivation of each of two kinds: from the axioms and the mathematical definitions alone when the target is a mathematical function, and from these together with the empirical definitions when it is an empirical function. The second kind of derivation is the one this paper is about, and it is the one needed by a hybrid AI in our sense where no verifier can check its output against the world.

As \citet{blum1996} has highlighted, there exists a profound conceptual tension between the continuous foundations of numerical analysis in machine learning (e.g., real-valued gradients) and the discrete foundations of computability theory (Turing machines). In this paper, we have taken a firm stance on the discrete nature of computation and demonstrated that learning true probabilities can be rigorously formalized as discrete syntactic deductions within $\mathcal{S}_t$ over computable reals. We have no finished answer to offer about the gap. We note only that our approach separates what is true of the world from what the machine can learn. The machine's learning, being computation, is discrete. If the world is in fact continuous, the mathematical consequences of a formal system that assumes continuity can still be used legitimately, provided that the machine's access to the consequences is by indices. The gap is produced by taking the discreteness of computation seriously, and it is not a flaw to be patched over. It remains a formidable challenge to fully reconcile the continuous mathematical models of machine learning with the discrete epistemic bounds of Turing computability. We regard it as the main open problem of the framework, and it is the subject of work now in progress.

Lastly, it is important to note that we have emphasized an epistemic conception of machine learning, rather than a simply computational one, in order to parallel machine learning with human learning. By the term ``machine'' in the ``machine learning'', we specifically refer to an artifact that is capable of performing behaviors at least at a human level of intelligence. Consequently, the epistemic dimension becomes essential for our analysis. As long as we aim to contribute to the field of artificial intelligence through the theory of machine learning, machines must be intelligent learners in this epistemic sense.

\paragraph{Acknowledgments.} Jinsook Kim was supported by Korea Institute for Advanced Study. Jinho Kang was supported by Gwangju Institute of Science and Technology.

\paragraph{AI Use Disclosure.} We used AI tools for research assistance, including discussing the arguments, editing sentences, and checking proofs, facts, and references. All material in the paper was written and verified by the authors.

\bibliographystyle{plainnat}
\bibliography{references}
   
\appendix

\section{Technical Proofs}
\label{app:proofs}

\subsection*{Proof of Theorem~\ref{thm:rep}}

Let $T$ be a formal system of arithmetic in the sense of Section~\ref{sec:theory}: it extends Robinson arithmetic $Q$, it is consistent, and its set of axioms and definitions is recursive. The system $\mathcal{S}_t$ is such a system at each $t$, since $\mathcal{A}$, $\mathcal{D}_{math}$ and $\mathcal{D}_{schema}$ are recursive and $\mathcal{D}_{fact}(t)$ is finite.

\paragraph{Step 1. Effective operations on indices via the $S^n_m$ theorem.}
An element $a \in \mathbb{R}_c^n$ is a vector of computable reals if for each $i$ there is a Turing machine that outputs a sequence of rationals converging to $a_i$ at the rate fixed in Section~\ref{sec:theory}. Let $e_{a_i}$ be the index of the machine provided to the learner for $a_i$. Let $f : \mathbb{R}_c^n \to \mathbb{R}_c$ be computable. Because $f$ is computable, there is an effective algorithm that takes the indices computing the inputs and returns an index computing the output $b = f(a)$. By the $S^n_m$ theorem \citep{kleene1952introduction}, this transformation is given by a partial recursive function $\psi$ on natural numbers. The index tracking function $h_f$ of $f$ is the restriction of $\psi$ to the valid indices of the arguments of $f$, on all of which $\psi$ halts because $f$ is computable. If the machine's run halts, the fact $h_f(e_{a_1}, \dots, e_{a_n}) = e_b$ holds in the metatheory, and $e_b$ is an index of $b$ because the machine computes $f$.

\paragraph{Step 2. Representability.}
By G\"odel's Representability Theorem \citep{Godel:31, Boolos:02}, every partial recursive function is representable in any consistent recursively axiomatized theory extending $Q$. Because $\psi$ is partial recursive over natural numbers and $T$ is such a theory, there is an arithmetical formula $\varphi_f(x_1, \dots, x_n, y)$ of $T$ representing $\psi$, and hence, for all valid indices $e_{a_1}, \dots, e_{a_n}$ and all numbers $e_b$,
\begin{equation}
\label{eq:repr}
h_f(e_{a_1}, \dots, e_{a_n}) = e_b \iff T \vdash \varphi_f(\overline{e_{a_1}}, \dots, \overline{e_{a_n}}, \overline{e_b}).
\end{equation}

\paragraph{Step 3. Conclusion.}
Equation \eqref{eq:repr} is the statement of the theorem: the left side says that the machine's run, given the indices of the arguments, returned the index $e_b$, and the right side says that $T$ derives the formula which represents that fact. So the machine's computation of $f$ on computable reals is matched, instance by instance, by a derivation in $T$ of the corresponding statement about indices. \qed

\subsection*{Proof of Theorem~\ref{thm:cond}}

Throughout, $\alpha^*$ is a fixed computable real in $[0,1]$ and $P(A_{t+1} \mid \mathcal{B}_t) = \alpha^*$ along the path considered. Write $\Pi_t = \Pi(A_{t+1} \mid \mathcal{B}_t)$, and let $c_t$ be the code of the record $\mathcal{D}_t$, so that the formula $\varphi_\Pi(\overline{c_t}, \overline{e_\alpha})$ represents, by Theorem~\ref{thm:rep}, the fact that the learner's index tracking function $h_\Pi$ sends the encoded record to the index $e_\alpha$.

\paragraph{Part I, the ``only if'' direction.}
Suppose that machines learn the true probability as $\alpha^*$.

\emph{The Truth Condition.} We consider two cases.

\emph{Case 1. $\alpha^*$ is a non-repeating irrational.}
Suppose the machine attempts to compute $\alpha^*$ extensionally, digit by digit. Let $S^{(i)}_t$ be the minimal formal system needed to compute and verify the expansion of $\alpha^*$ up to the $i$-th place at time $t$. Because $\alpha^*$ is non-repeating, each further digit requires information not derivable from the previous finite digits, so the systems form a strictly increasing chain $S^{(1)}_t \subsetneq S^{(2)}_t \subsetneq \cdots$. To complete the extensional computation, the machine would have to operate inside the inductive limit $S^{(\infty)}_t = \bigcup_{i \ge 1} S^{(i)}_t$. By Turing computability and by classical proof theory \citep{turing1939systems, feferman1962transfinite}, a valid formal proof is a finite sequence of symbols verifiable in finite time. The system $S^{(\infty)}_t$ is not finitely axiomatizable, so the machine can never mechanically complete the extensional computation.

The machine must therefore abandon extensional evaluation and adopt an intensional approach, using a fixed finitely axiomatizable system $S \subseteq S^{(i)}_t$ for all $i$. This system encodes the finite algorithm $h_\Pi$ that generates $\alpha^*$. Inside $S$ the machine does not compute infinite digits. It uses the universal quantifier to derive, in finitely many steps, the arithmetical formula \eqref{eq:conv} of Section~\ref{sec:true}. All quantifiers in \eqref{eq:conv} range over natural numbers, and no analytic limit is used. Since $S$ exists at some finite $t_0$, and $S \subseteq S^{(i)}_t$ for all $t > t_0$, the index tracking function $h_\Pi$ generates $\alpha^*$, so by Theorem~\ref{thm:rep} and Definition~\ref{def:subj} there is a subjective probability $\Pi_t$ computed at each stage $t > t_0$. In the metatheory, \eqref{eq:conv} gives $|\Pi_t - \alpha^*| \le \tpw{k}$ for all $t \ge N$ and every $k$ (Appendix~\ref{app:remarks}), hence
\[
\limsup_{t \to \infty} |\Pi_t - P(A_{t+1} \mid \mathcal{B}_t)| < \epsilon \quad \text{for every } \epsilon > 0.
\]

\emph{Case 2. $\alpha^*$ is rational.}
Then $\alpha^*$ is represented by finitely many digits at a finite time $t$. Because the machine learns the true probability, by Definition~\ref{def:learn} and Definition~\ref{def:success} it successfully computes $\alpha^*$ by a proper derivation, so there is a system $\mathcal{S}_t = \mathcal{A} \cup \mathcal{D}_{math} \cup \mathcal{D}_{emp}(t)$ with
\[
\mathcal{S}_t \vdash \varphi_\Pi(\overline{c_t}, \overline{e_{\alpha^*}}),
\]
the formula saying that the learner, given the record, returned an index of $\alpha^*$. Suppose the true probability takes the value $\alpha^*$ along the path without limit, and let $\{\tau_k\}_{k \ge 1}$ be the subsequence of times at which it does. Along this subsequence the time average of the indicator $Y_{\tau_k + 1} = \mathbf{1}\{A_{\tau_k+1}\}$ coincides with $\alpha^*$. This is ergodicity in an extended sense. No stationarity is assumed of the process, and the coincidence of the time average with the value holds along the subsequence on which the conditional law takes that value. Because the machine computes $\Pi_{\tau_k} = \alpha^*$ at these instances, we get $|\Pi_{\tau_k} - P(A_{\tau_k+1} \mid \mathcal{B}_{\tau_k})| = 0$, which yields the Truth Condition.

\emph{The Assurance Condition.} Given the Truth Condition, there is a stochastic path on which $\Pi_t = \alpha^* = P(A_{t+1} \mid \mathcal{B}_t)$ for all $t > t_0$. Put
\[
\xi_{t+1} = \begin{cases} 1 & \text{if } \Pi_t = \alpha^* = P(A_{t+1} \mid \mathcal{B}_t) \\ 0 & \text{if } \Pi_t = \alpha^* \ne P(A_{t+1} \mid \mathcal{B}_t) \end{cases}
\qquad
M_t = \Bigl( \sum_{j=1}^{t} \xi_j \Bigr)^{-1} \xi_t (Y_t - \alpha^*),
\]
where $Y_{t+1}$ is the indicator of $A_{t+1}$. Along the path, $\Pi_t - \alpha^* = \mathbb{E}[Y_{t+1} \mid \mathcal{B}_t] - \alpha^* = 0$ for all $t > t_0$. Both $(\sum_{j \le t+1} \xi_j)^{-1}$ and $\xi_{t+1}$ are $\mathcal{B}_t$-measurable, and $\mathbb{E}[Y_{t+1} - \alpha^* \mid \mathcal{B}_t] = 0$ along the path, so $\mathbb{E}[M_{t+1} \mid \mathcal{B}_t] = 0$, where the expectation is taken with respect to $\Pi(\cdot \mid \mathcal{B}_t)$. Hence $\sum_{t=t_0}^{k} M_t$ is a martingale adapted to $\mathcal{B}_{k-1}$ along the path. Moreover
\[
\mathbb{E}\Bigl[ \Bigl( \sum_{t=1}^{k} M_t \Bigr)^2 \Bigr] = \sum_{t=1}^{k} \mathbb{E}[M_t^2] \le \lambda \sum_{t=1}^{k} \mathbb{E}\Bigl[ \Bigl( \bigl( \textstyle\sum_{j \le t} \xi_j \bigr)^{-1} \xi_t \Bigr)^2 \Bigr] \le \frac{\lambda \pi^2}{6},
\]
because $Y_t$ is an indicator variable, so that $\mathrm{var}(Y_t \mid \mathcal{B}_t)$ is bounded above uniformly by some $\lambda$ with $0 \le \lambda < \infty$. By the martingale convergence theorem and Kronecker's lemma,
\[
\Bigl( \sum_{t=0}^{k-1} \xi_{t+1} \Bigr)^{-1} \sum_{t=0}^{k-1} \xi_{t+1} (Y_{t+1} - \alpha^*) \longrightarrow 0 \quad \text{with } \Pi\text{-probability one}.
\]
By the law of iterated expectation and fixing $P(A_{t+1} \mid \mathcal{B}_t) = \alpha^*$, we obtain the Assurance Condition.

\paragraph{Part II, the ``if'' direction.}
Suppose both conditions hold. From the Truth Condition, $\Pi_t$ converges to $\alpha^*$. The Assurance Condition assigns subjective probability exactly $1$ to that infinite convergence. Because the machine is a finite device, we must ask how it can assign certainty to an infinite sequence of events. By \citet{turing1939systems} and \citet{feferman1962transfinite}, an infinite inductive limit of formal systems cannot be mechanically verified in finite time, so a Turing machine cannot assign $\Pi = 1$ by checking infinitely many future samples extensionally.

The only permissible mechanism is a finite syntactic derivation inside $\mathcal{S}_t$. That is, there is a finitely axiomatizable $S \subseteq \mathcal{S}_t$ for all $t > t_0$ in which the machine derives formula \eqref{eq:conv}. Call that formula $\phi$. By Definition~\ref{def:success}, the machine operates within $\mathcal{A} \cup \mathcal{D}_{math} \cup \mathcal{D}_{emp}(t)$. Because the target is an empirical true probability and not a purely mathematical object, the derivation cannot proceed from logical and mathematical resources alone. It requires the empirical definitions grounded in the record filtration. That is,
\[
\mathcal{A} \cup \mathcal{D}_{math} \nvdash \phi \quad \text{but} \quad \mathcal{A} \cup \mathcal{D}_{math} \cup \mathcal{D}_{emp}(t) \vdash \phi.
\]
Because $\phi$ is derived within $\mathcal{S}_t$ and the Truth Condition guarantees $P(A_{t+1} \mid \mathcal{B}_t) = \alpha^*$ along the path that $\phi$ describes, the formula $\varphi_{P}(\overline{c_t}, \overline{e_{\alpha^*}})$ is also derived within $\mathcal{S}_t$, where $\varphi_P$ represents the index tracking function of the target $P(A_{t+1} \mid \mathcal{B}_t)$; that is, the value $\alpha^*$ returned on the record is derived to be the value of the target. The machine therefore achieves successful computation of $\alpha^*$ by Definition~\ref{def:success}, and by Definition~\ref{def:learn} it learns the true probability as $\alpha^*$. \qed

\subsection*{Proof of Theorem~\ref{thm:approx}}

We argue by contradiction. Assume (A1) that no ideal machine can learn $f^*$, and (A2) that for every $\delta_0 > 0$ there are $\delta < \delta_0$ and a finite learning rule $f_{n,m}$ such that a real machine successfully computes that $f_{n,m}$ lies within $\delta$ of $f^{\ast}$. Here an ideal machine is a classical Turing machine with no bound on time and no bound on tape.

By (A2), for each $k$ there is a finite rule $f_{n_k, m_k}$ and a proper derivation in $\mathcal{S}_t$ that it lies within $\tpw{k}$ of $f^{\ast}$. Whether a string is a derivation in $\mathcal{S}_t$ is decided by a mechanical check, so an ideal machine can find these derivations: given $k$, it enumerates strings until it finds such a derivation, and then runs the rule it has found. So for each $k$ the ideal machine outputs an approximation to $f^{\ast}$ within $\tpw{k}$, and each of these outputs has a proper derivation. Since the outputs of a machine computing a real-valued function are its approximations at the requested precisions (Section~\ref{sec:theory}), this is successful computation of $f^{\ast}$ instance by instance.

By Definition~\ref{def:success}, this means that $f^*$ is successfully computable by the ideal machine, which contradicts (A1). Hence (A2) is false, which is the statement of the theorem. \qed

\subsection*{Proof of Theorem~\ref{thm:tolerate}}

Suppose the machine satisfies the Truth Condition but not the Assurance Condition. Then
\[
\Pi\bigl( \Pi_t = \alpha \ne P(A_{t+1} \mid \mathcal{B}_t) \text{ infinitely often} \bigr) > 0 .
\]
It follows that $\liminf_{t} \Pi(\Pi_t = \alpha \ne P(A_{t+1} \mid \mathcal{B}_t)) \ge \Pi(\Pi_t = \alpha = P(A_{t+1} \mid \mathcal{B}_t) \text{ finitely often}) > 0$. Hence $\Pi(P(A_{t+1} \mid \mathcal{B}_t) \ne \alpha) > 0$ for all but finitely many $t$, which by Definition~\ref{def:tolerate} means that the machine tolerates the possibility of error at all but finitely many of its infinitely many opportunities to learn, since $P(A_{t+1} \mid \mathcal{B}_t) = \alpha = \Pi_t$ by the Truth Condition. \qed

\subsection*{Proof of Theorem~\ref{thm:circ}}

The proof shows the loop directly. To keep the notation apart from the conditional probabilities, we write $Z$ for the instance space of the PAC problem and $z$ for its elements. The letter $X$ is reserved for the process.

\paragraph{The PAC setting.}
A concept over $Z$ is a subset $c \subseteq Z$, equivalently a boolean function $f : Z \to \{0,1\}$. Let $D$ be a fixed distribution over $Z$. For a hypothesis $g$,
\begin{equation}
\label{eq:err}
\mathrm{err}(g) = \Pr_{z \sim D}\bigl[ g(z) \ne f(z) \bigr].
\end{equation}
The concept $f$ is PAC learnable by $g$ if an algorithm receives $m(\epsilon, \delta)$ samples drawn i.i.d.\ from $D$, with $m$ polynomial in $1/\epsilon$ and $1/\delta$, and outputs $g$ with $\Pr[\mathrm{err}(g) \le \epsilon] \ge 1 - \delta$.

\paragraph{The meta-learning target.}
Let $X = (X_t)_{t}$ be a process with a discrete state. In our setting the object to be learned is the conditional law itself, so the value $P(X_t = x \mid \mathcal{B}_{t-1})$ plays the role of the realized sample. Let $V$ be the set of values this conditional probability takes. We take $V = \{\alpha, \beta\}$ with $\alpha \ne \beta$ for readability, and any finite or countable $V$ would do without change to the argument. Let $Z^*_\alpha$ be the set of stages on which the conditional probability equals $\alpha$ and let $Z^*_\beta$ be the set of stages on which it equals $\beta$. Define
\[
f(z) = \begin{cases} 1 & \text{if } z \in Z^*_\alpha \\ 0 & \text{if } z \in Z^*_\beta \end{cases}
\qquad
g(z) = \begin{cases} 1 & \text{if } \Pi(X_t = x \mid \mathcal{B}_{t-1}) = \alpha \text{ at } z \\ 0 & \text{otherwise.} \end{cases}
\]
The hypothesis $g$ is computed by the machine from its own estimate $\Pi$ alone.

\paragraph{The loop.}
We establish $A \Rightarrow B \Rightarrow C \Leftrightarrow A$, where $A$ is the assumption that the samples of the true probability are i.i.d., $B$ is that $f$ is PAC learnable, and $C$ is that the machine can compute the error \eqref{eq:err}.

$A \Rightarrow B$ holds by the premise of the PAC framework. $B \Rightarrow C$ holds because a PAC learnable concept has a computable error by definition.

For $C \Leftrightarrow A$, note that for the samples to be identically distributed, the probability that the true conditional probability takes a given value must be invariant across stages. If the machine assumes identical distribution, then it assumes that its samples come exclusively from one of $Z^*_\alpha$ and $Z^*_\beta$. Without loss of generality let it assume $Z^*_\alpha$, so that $\Pi$ is fixed at $\alpha$. Then along $Z^*_\alpha$ we get $f = 1 = g$ on the stages where the conditional probability is $\alpha$, and $f = 0 = g$ elsewhere. Along $Z^*_\beta$ we get $f = 0$ while $g = 1$ on those same stages. Hence
\[
g(z) = f(z) \iff \Pi(X_t = x \mid \mathcal{B}_{t-1}) = \alpha = P(X_t = x \mid \mathcal{B}_{t-1}).
\]
So computing the PAC error $\Pr[g(z) \ne f(z)]$ is the same task as computing the probability that the true conditional probability deviates from $\alpha$. The ability to compute the error is therefore equivalent to the assumption that the true probabilities are identically distributed, which is $A$.

\paragraph{Where the loop bites.}
The idea the proof turns on can be stated without the counterexample. The error \eqref{eq:err} is defined relative to a fixed $D$, and the sample complexity bound requires i.i.d.\ sampling from $D$. When the target concept is the conditional law of the data-generating process, $D$ is the law of the values the target takes, so the target determines it. So the criterion by which learning is measured presupposes the object to be learned. By Definition~\ref{def:circ} this is epistemic circularity. Note that the chain $A \Rightarrow B \Rightarrow C$ uses no sentence of $\mathcal{D}_{fact}(t)$, so the certificate $C$ is derivable from $\mathcal{A} \cup \mathcal{D}_{math}$ together with the assumption $A$ alone, which is the condition of Definition~\ref{def:mere}. This is no accident. Since $C$ is equivalent to the premise $A$, any derivation of $C$ is a derivation from $A$, and the record is necessarily idle in it. Circularity entails that the data do no work, and by Definition~\ref{def:mere} what the machine performs is a mere computation. \qed

\subsection*{Proof of Theorem~\ref{thm:corpus}}

\paragraph{Step 1. The ideal corpus is finite.}
Let each word $w$ of length $n_w$ be a sequence of letters and let each sentence $S$ of length $n_s$ be a sequence of words that humans have circulated in order to communicate up to $t_0$. A sentence is distinguished from an arbitrary ungrammatical string that cannot be used for communication. For any $w \in W^*$ the length $n_w$ is finite, since no word whose sequence goes to infinity can be used. Likewise $n_s$ is finite for any $S \in W^*$. Let $n^*_w$ and $n^*_s$ be the maximum lengths in $W^*$. The population of possible combinations $W_{\max}$ has size bounded by $\lambda \cdot 26^{(n^*_w \times n^*_s)}$ for some countable $\lambda$, where $\lambda$ is countable because at most countably many people have used the language up to $t_0$ in a world where time is discrete. At any $t_0$ there is a set $W^* \subseteq W_{\max}$ containing the accumulated history of circulated words and sentences, and the size of $W^*$ is finite. The alphabet may be extended to numerals and punctuation without change to the argument.

\paragraph{Step 2. The true probability is computable from the corpus.}
Define a counting function $f_c$ from $W^*$ to $\mathbb{N}$, counting the frequency of each sentence. Because the set of strings is lexicographically Turing-enumerable and $W^*$ is finite, $f_c$ is computable. With $P(S \mid \mathcal{B}_t) = P(w^n_1 \mid \mathcal{B}_t) = \prod_{k=1}^{n} P(w_k \mid w_1^{k-1}, \mathcal{B}_t)$, the machine computes
\[
\Pi(w^n_1 \mid \mathcal{B}_t) = \prod_{k=1}^{n} \frac{C(w_k, w_{k-1}, \dots, w_1)}{C(w_{k-1}, \dots, w_1)},
\]
where $C(\cdot)$ are the counts given by $f_c$ and $C(w_0)$ is the number of all sentences in the corpus.

\paragraph{Step 3. From computation to learning.}
Since $f_c$ is derived from $W^*$, the value of $\Pi$ so computed is the true probability of Definition~\ref{def:trueprob}. Because $f_c$ is computable over the finite domain $W^*$, Theorem~\ref{thm:rep} guarantees that the derivation of $\Pi$ is a formal derivation inside $\mathcal{S}_t$, and it draws on the record of $W^*$ in $\mathcal{D}_{emp}(t)$, so the process is a successful computation under Definition~\ref{def:success}. By Definition~\ref{def:learn} the true probability is learned. \qed

\medskip
\noindent \emph{Proof of Proposition~\ref{prop:nonvac}.} Immediate from Theorem~\ref{thm:corpus}, since a target that is learned satisfies the conditions of Theorem~\ref{thm:cond} by the ``only if'' direction of that theorem. \qed

\section{Further Remarks}
\label{app:remarks}

\begin{remark}[Non-uniqueness of indices]
\label{rem:B1}
Infinitely many Turing machines compute the same real $a$, so the map from reals to indices is one to many. The ambiguity does not enter Theorem~\ref{thm:rep}, because the theorem concerns the index tracking function $h_f$ of the machine that was actually run: the derivation is about the specific computation path $\overline{e_a} \mapsto \overline{e_b}$ that the machine took, and not about the real $a$ as such.
\end{remark}

\begin{remark}[The two levels]
\label{rem:B2}
Three roles are kept apart in Section~\ref{sec:true}. The learner is the machine of Definition~\ref{def:subj}: at each stage $t$ it reads the record on its tape and outputs a natural number $e_t$, the index of a program. The program is the second machine: given a precision request $n$, it prints a rational $q_t(n)$ within $\tpw{n}$ of the real it computes, which is $\Pi_t$. The statement that the values $\Pi_t$ converge to $\alpha^*$ in $\mathbb{R}_c$ is ours, made in the metatheory. What the learner derives inside $\mathcal{S}_t$ is formula \eqref{eq:conv}, and we set out here what that formula says and why it is convergence.

\emph{Kleene normal form.} In \eqref{eq:conv}, $q_t(n) = q$ abbreviates $\exists s\,(T(e_t, n, s) \wedge U(s) = \bar{c})$, where $T$ is Kleene's $T$-predicate, $U$ the output function, and $c$ the code of the rational $q$; both are primitive recursive, so the abbreviation is an arithmetical formula. From some stage $t_0$ on, the indices the learner outputs are determined by its fixed algorithm $h_\Pi$ and the record, so the map $t \mapsto e_t$ is arithmetically definable and \eqref{eq:conv} is a single formula about it. All of its quantifiers range over natural numbers, and no limit enters as an object.

\emph{Why \eqref{eq:conv} is convergence.} Fix $t$ and $k$. If $|\Pi_t - \alpha^*| \le \tpw{k}$, then for every $n$,
$$ |q_t(n) - q_{\alpha^*}(n)| \le |q_t(n) - \Pi_t| + |\Pi_t - \alpha^*| + |\alpha^* - q_{\alpha^*}(n)| \le \pw{k} + \pw{n-1}, $$
by the triangle inequality and the rate fixed in Section~\ref{sec:theory}, which bounds the first and the third term by $\tpw{n}$ each. Conversely, if $|q_t(n) - q_{\alpha^*}(n)| \le \tpw{k} + \tpw{n-1}$ for every $n$, the same inequality, read the other way, gives $|\Pi_t - \alpha^*| \le \tpw{k} + \tpw{n-2}$ for every $n$, and since the left side does not depend on $n$, $|\Pi_t - \alpha^*| \le \tpw{k}$. So the part of \eqref{eq:conv} after $\forall t \ge N$ is equivalent to $|\Pi_t - \alpha^*| \le \tpw{k}$, and \eqref{eq:conv} is equivalent to the convergence of $\Pi_t$ to $\alpha^*$. The equivalence uses only the rate of the two programs, so no constant has to be chosen.

\emph{The temporal paradox.} Assigning probability $1$ at a finite stage to an infinite future convergence may look like a temporal paradox. At stage $t$ the machine has read finitely many records, and \eqref{eq:conv} quantifies over stages it has not yet seen. The resolution is that what the machine has is not a completed infinite computation but a finite derivation. A finite derivation can speak of infinitely many instances by means of the universal quantifier, and it establishes the formula for every instance without surveying the instances as a completed whole. What would be a paradox for computation is routine for proof. What the derivation requires, and what finite data can supply, is another matter: Theorem~\ref{thm:cond} says when the derivation is available, and Theorem~\ref{thm:corpus} gives a domain in which it is.
\end{remark}

\begin{remark}[The syntactic form of the empirical definitions]
\label{rem:B3}
Each sentence of $\mathcal{D}_{schema}$ introduces a function symbol $\mathsf{X}^{(j)}_s$ by a condition $\delta_j$ that is primitive recursive and for which $\mathcal{A}$ proves $\forall x\, \exists!\, y\, \delta_j(x,y)$, so the symbol denotes a total single-valued function on codes. Each sentence of $\mathcal{D}_{fact}(t)$ is an atomic equation $d^{(j)}_k = \bar n$ in a name new to the language. Both are definitional extensions of $\mathcal{A} \cup \mathcal{D}_{math}$ in the usual sense: every formula of the old language that is derivable in $\mathcal{S}_t$ is already derivable in $\mathcal{A} \cup \mathcal{D}_{math}$, so the empirical definitions introduce no new theorem about numbers and, in particular, no inconsistency.
\end{remark}

\begin{remark}[The observables separate the states]
\label{rem:B4}
Section~\ref{sec:true} takes the family of observables $(X^{(j)}_t)_{j \in J}$ to separate the states of $\Omega_t$: for any two distinct states some observable takes different values on them. Since $\Omega_t$ is enumerable, every singleton $\{\omega_i\}$ is then an intersection of countably many sets of the form $\{X^{(j)}_t = x\}$, so every singleton, and hence every subset of $\Omega_t$, belongs to the $\sigma$-algebra generated by the observables at time $t$. This is why $\mathcal{B}_t$ is the totality of facts up to time $t$ and not only the facts visible at the resolution of the observables. Without the separation condition, $\mathcal{B}_t$ would record the facts only up to that resolution: if, for example, the states were (rain, wind), (rain, calm) and (sun, calm) and the only observable were the indicator of rain, the first two states would not be separated and the fact of wind would not be in $\mathcal{B}_t$.
\end{remark}

\begin{remark}[The two questions about sentences]
\label{rem:B5}
The Markov question and our question differ, and the difference matters because human languages are productive. A new sentence that has not been circulated has zero frequency in the ideal corpus, however large the corpus is set. So the probability of obtaining an arbitrary new sentence would be zero if computed from the empirical distribution in the corpus. That cannot be right if the language is productive. So the probability of obtaining a new sentence cannot be defined by its frequency in $W^*$, and Theorem~\ref{thm:corpus} does not cover it. We state what is missing as a problem.

\begin{openproblem}[Productivity]
\label{op:productivity}
Let $S$ be a sentence with zero frequency in $W^*$. Find a definition of the true probability of obtaining $S$, and a formal system $\mathcal{S}_t$, such that (a) the definition agrees with $P(S' \mid \mathcal{B}_t)$ for every sentence $S'$ in $W^*$, and (b) the probability of $S$ is learned by a machine in the sense of Definition~\ref{def:success}, that is, derived properly in $\mathcal{S}_t$ and not assumed.
\end{openproblem}

Theorem~\ref{thm:corpus} solves the problem for sentences in $W^*$. There the record decides the probability. For a sentence outside $W^*$ no finite record decides it, so condition (b) cannot be met by $\mathcal{D}_{emp}(t)$ alone. In the hybrid framework of this paper, what the proposer lacks for such a sentence is a verifier: a decidable check that closes each new sentence as the proof relation $\Prf$ closes each candidate proof. For natural language no such verifier is known, since it would require a recursive grammar and a theory of meaning inside $\mathcal{S}_t$.

This zero-frequency problem does not arise for speech recognition on a simple $N$-gram model, where inputs are given. The unambiguous form of any input given to the machine must already be in the ideal corpus, since the input must actually have been used in speech before it was given to the machine. Even in the extreme case where the input was created at the moment the speech was made, the true probability of obtaining that sentence is not zero.
\end{remark}


\end{document}